\documentclass{bmvc2k}
\usepackage{times}
\usepackage{epsfig}
\usepackage{graphicx}
\usepackage{amsmath}
\usepackage{amssymb}
\usepackage{index}
\usepackage{color}
\usepackage{array}
\usepackage{multirow}
\usepackage{rotating}
\usepackage{epstopdf}
\usepackage{algpseudocode}
\usepackage{algorithm}
\usepackage{setspace}
\usepackage{eqnarray}
\usepackage{subfigure}
\usepackage{breqn}
\usepackage{capt-of}
\usepackage{wrapfig}
\usepackage{comment}
\usepackage[table]{xcolor}

\title{WxBS: Wide Baseline Stereo Generalizations}

\addauthor{Dmytro Mishkin}{ducha.aiki@gmail.com}{1}
\addauthor{Jiri Matas}{matas@cmp.felk.cvut.cz}{1}
\addauthor{Michal Perdoch}{perdom1@cmp.felk.cvut.cz}{1}
\addauthor{Karel Lenc}{karel@robots.ox.ac.uk}{2}

\addinstitution{
 Center for Machine Perception\\
 Czech Technical University in Prague\\
 Czech Republic
}
\addinstitution{
Visual Geometry Group\\
Department of Engineering Science\\
University of Oxford\\
Oxford, UK
}

\runninghead{Mishkin et al.}{WxBS: Wide Baseline Stereo Generalizations}

\newcommand{\algrule}[1][.2pt]{\par\vskip.5\baselineskip\hrule height #1\par\vskip.5\baselineskip}

\newenvironment{itemize*}%
  {\begin{itemize}%
    \setlength{\itemsep}{1pt}%
    \setlength{\parskip}{1pt}}%
  {\end{itemize}}

\newenvironment{enumerate*}%
  {\begin{enumerate}%
    \setlength{\itemsep}{1pt}%
    \setlength{\parskip}{1pt}}%
  {\end{enumerate}}

\graphicspath{{./figures/}{./figures/pairs_gt_err/}}

\newindex{todo}{todo}{tnd}{Todo List}

\newcommand{\executeiffilenewer}[3]{%
\ifnum\pdfstrcmp{\pdffilemoddate{#1}}%
{\pdffilemoddate{#2}}>0%
{\immediate\write18{#3}}\fi%
}

\makeatletter
\renewcommand{\paragraph}{%
  \@startsection{paragraph}{4}%
  {\z@}{0.5ex \@plus 1ex \@minus .2ex}{-1em}%
  {\normalfont\normalsize\bfseries}%
}

\def\eg{\emph{e.g}\bmvaOneDot}
\def\ie{\emph{i.e.}\bmvaOneDot}

\def\etal{\emph{et al}\bmvaOneDot}

\def\ccaEF#1{\cellcolor{black!#1!black!#1}\ifnum #1>50\color{white}\fi{#1}}
\def\ccaEVD#1{\cellcolor{black!#10}\ifnum #1>7\color{white}\fi{#1}}
\def\ccaMMS#1{\cellcolor{black!#1}\ifnum #1>50\color{white}\fi{#1}}
\def\ccaWGABS#1{\cellcolor{black!#10}\ifnum #1>3\color{white}\fi{#1}}
\def\ccaWGALBS#1{\cellcolor{black!#10}\ifnum #1>5\color{white}\fi{#1}}
\def\ccaWGLBS#1{\cellcolor{black!#10}\ifnum #1>5\color{white}\fi{#1}}
\def\ccaWGSBS#1{\cellcolor{black!#10}\ifnum #1>3\color{white}\fi{#1}}
\def\ccaWLABS#1{\cellcolor{black!#10}\ifnum #1>3\color{white}\fi{#1}}
\def\ccalostinpast#1{\cellcolor{black!#1}\ifnum #1>80\color{white}\fi{#1}}
\def\ccaoxford#1{\cellcolor{black!#1!black!#1}\ifnum #1>50\color{white}\fi{#1}}
\def\ccasnavely#1{\cellcolor{black!#1!black!#1}\ifnum #1>50\color{white}\fi{#1}}
\def\ccastewart#1{\cellcolor{black!#1}\ifnum #1>55\color{white}\fi{#1}}

\begin{document}

\maketitle

\begin{abstract}
We present a generalization of the wide baseline two view matching problem -
\textsc{WxBS}, where \textsc{x} stands for a different subset of ``wide baselines" in acquisition conditions such as geometry, illumination, sensor and appearance.
We introduce  a novel dataset of ground-truthed image pairs which include multiple "wide baselines" and show that state-of-the-art matchers fail on almost all image pairs from the set.
A novel matching algorithm for addressing the \textsc{WxBS} problem is introduced and we show experimentally that the \textsc{WxBS-M} matcher dominates the state-of-the-art methods both on the new and existing datasets. 
\end{abstract}

\section{Introduction}
\label{sec:introduction}
%
\begin{wrapfigure}{r}{0.5\textwidth}
\begin{center}
\def\svgwidth{1\textwidth}
\includegraphics[width=0.45\textwidth]{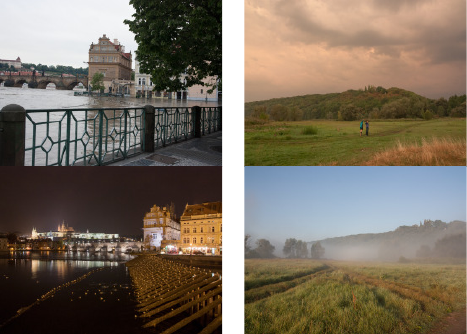}
\end{center}
\caption{Examples of \textsc{WxBS} problems.}
\label{fig:wxbs-cover}
\vspace{-1em}
\end{wrapfigure}
The Wide Baseline Stereo (\textsc{WBS}) matching problem, first formulated by Pritchett and Zisserman~\cite{Pritchett1998}, has received significant attention in the last 15 years \cite{Mikolajczyk2005,Tuytelaars2008}. Progressively more challenging two- and multi-view problems have been successfully handled \cite{Tuytelaars2008} and recent algorithms~\cite{Morel2009}, \cite{Mishkin2015} have shown impressive performance, e.g. matching views of planar objects with orientation difference of up to 160 degrees.

Besides the orientation and viewpoint baseline, other factors influence the complexity of establishing geometric correspondence between a pair of images. The standard physical models of image formation and acquisition consider, beside geometry,  the effects of illumination, the properties of the transparent medium light rays pass through in the scene, the surface properties of objects and the properties of the imaging sensors.

In the paper, we consider the generalization of Wide (geometric) Baseline Stereo to \textsc{WxBS}, a two-view image matching problem where two or more of the image formation and acquisition properties significantly change, i.e. they have a {\it  wide baseline}. The  "significant change" distinguishes the problem from image registration, where dense correspondence is routinely established between multi-modal images and
various complex transformations have been considered, see Zitov{\'a} and Flusser~\cite{Zitova2003}. Operationally, the "wide baseline" means "where local, gradient-descent type" methods fail.

The following single wide baseline stereo, or correspondence, problems and their combinations are considered: illumination (\textsc{WlBS}) -- difference in position, direction, number, intensity and wavelength of light sources; geometry  (\textsc{WgBS}) -- difference in camera and object pose, scale and resolution - the ``classical'' WBS;
sensor  (\textsc{WsBS}) -- change in sensor type: visible, IR, MR; noise, image preprocessing algorithms inside the camera, etc;
appearance (\textsc{WaBS}) --  difference in the object appearance because of time or seasonal changes, occlusions, turbulent air,  etc.
We denote matching problems, or, equivalently, image pairs, with a significant change in only one of the groups listed  as {W{\sc1}BS}; if a combination of effects is present, as \textsc{WxBS}. To our knowledge, almost all published image datasets and algorithms are in the 
\textsc{W1BS} class\cite{Mikolajczyk2005}, \cite{Morel2009}, \cite{Vonikakis2013},\cite{Aguilera2012},\cite{Hauagge2012}, \cite{Jacobs2007}. 

We present a new public dataset with ground truth which combines the above-mentioned challenges and contains both \textsc{W2BS} image pairs  including viewpoint and appearance, viewpoint and illumination, viewpoint and sensor, illumination and appearance change and \textsc{W3BS} -- problems where viewpoint, appearance and lighting differ significantly. 

We show that state-of-the-art matchers performs poorly on the introduced image matching pairs, and propose a novel algorithm 
which significantly outperforms the state-of-the-art without a dramatic loss of speed. 

The paper is organised as follows. In Section~\ref{sec:related_work}, relevant datasets and matching algorithms are reviewed. The novel
\textsc{WxBS} matching algorithm is then introduced in Section~\ref{sec:wxbs-matcher}.  The dataset for \textsc{WxBS} problems and the associated evaluation protocol are presented in Section~\ref{sec:protocol}. Experimental results are described in Section~\ref{sec:results}. The paper is concluded in Section~\ref{sec:conclusions}.

\section{Related Work}
\label{sec:related_work}

\noindent
{\bf Viewpoint change.} The stereo problem -- matching of two images taken from different viewpoints --  has always received significant attention of the computer vision community as it is a critical component of the structure from motion task. For images taken concurrently, in both the calibrated and uncalibrated  set up, the problem for a narrow baseline is mature \cite{Tuytelaars2008} and can be now solved in real-time and on a large scale \cite{Agarwal2009}.

For wide-baseline matching, the standard evaluation protocol focuses on the feature detection and description stages\cite{Mikolajczyk2005}. However, the methodology and datasets of \cite{Mikolajczyk2005} are limited to images related by a homography. Attempts have been made to extend the evaluation to 3D scenes~\cite{Moreels2005, Aanaes2012}, but they are significantly less popular.
Neither of the above-mentioned protocols evaluates the performance of the matching stage and thus of the full matching pipeline.

As a reference, we adopted two recent algorithms which reported good performance and whose binaries are freely available.
The ASIFT method \cite{Morel2009} method synthetically transforms images in order to improve the range of  affine transformations of the DoG detector. This idea have been further extended in MODS~\cite{Mishkin2013} which incorporates multiple detectors and adopts an iterative approach that attempts to minimize the matching time. Both algorithms are able to match images with extreme viewpoint changes. Mishkin \etal~\cite{Mishkin2013} introduced an extreme-viewpoint dataset that is used to test the ability of the newly proposed WxBS matcher to handle viewpoint changes.

\noindent
{\bf Multimodal image analysis} is needed for the alignment of images acquired by different sensors. Most commonly, the problem is encountered in remote sensing and in medical imaging. For instance, in  \cite{Ghassabi2013}, \emph{red-free}  and \emph{fluorescein angiographic} images are matched. Similarly for different modes of magnetic resonance imaging, modality of the captured data depends on the magnetic properties of the scanned chemical compound.  In remote sensing, multimodal matching involves, \eg registering visual spectrum images against near infrared images (NIR) or Long-Wave infrared (LWIR).

Multimodal registration methods are usually divided to area-based and feature-based methods. As we are interested in extending the challenges into multiple-baseline variations, area-based methods are omitted as they lack scale invariance~\cite{Ghassabi2013}.

Feature-based approaches~\cite{Vonikakis2013} and~\cite{Ghassabi2013} identify the main issues of existing algorithms in the context of multimodal matching as the selection of the the response threshold, i.e.  the minimal image contrast which triggers the detector. In~\cite{Vonikakis2013}, the Difference of Gaussian (DoG)  \cite{Lowe2004} response is normalised by local average image intensity in cases when the image contrast is low.  Ghassabi \etal~\cite{Ghassabi2013} present a variant of the DoG detector which sets a local response threshold for each image cell on the basis of the image entropy. In~\cite{Chen2010}, it is argued that Harris detector is more suitable for this task as the information along boundaries is preserved in cases of different image modalities.

The main issue of the widely used SIFT descriptor~\cite{Lowe2004} in the context of multimodal images  is the lack of invariance to gradient reversal. Two approaches to address this issue have been proposed in the literature.
The first generates a second SIFT descriptor of the feature for a gradient reversed image by SIFT vector reordering~\cite{Hare2011}. We refer to this method as inverted-SIFT.  The second method~\cite{Chen2010}, denoted as half-SIFT, limits local image gradients directions to $\langle 0, \pi )$ by merging opposite gradient directions in orientation estimation. Unlike the inverted-SIFT, this method allows matching of images that are only partially inverted (per patch),\ie some gradient directions stay the same while other are reversed. The downside is the reduction of the descriptor discriminability. 

The computation of inverted-SIFT has a negligible computational cost, as it can be generated from SIFT descriptors by rearranging the data in the gradient histogram. The only associated computational cost is in the matching since twice as many features are matched in the second image. 
For the half-SIFT method, the feature patch and its descriptor has to be extracted as the dominant feature orientation differs from SIFT's dominant orientation.

An example of a multimodal image registration dataset is presented in \cite{Aguilera2012}. This dataset consist of 100 pairs of vertically aligned images from a camera and a LWIR thermal sensor. The viewpoint changes between related image pairs are negligible.

\noindent
{\bf Change in object illumination and appearance.}
Techniques similar to those developed for multimodal image matching can be used for matching of images of differently illuminated objects. 
In \cite{Kelman2007}, the authors employ half-SIFT and further modify SIFT descriptor in such a way that it collects only gradients located on edges. Yang \etal~\cite{Yang2007}  use the Difference of Gaussian features and SIFT to estimate the transformation between the images. If no matches are found, an identity transformation is assumed. From a single local match, multiscale features together with local image statistics are used in an iterative procedure called Dual-Bootstrap to enlarge the region of good alignment. 
A data presented in \cite{Kelman2007} are used in  Section~\ref{sec:results}. 

Hauagge \etal~\cite{Hauagge2012} argue that local symmetries survive significant illumination changes and developed a higher-level feature detector for matching of urban scenes where symmetries are abundant. They also assume that the vertical direction is aligned with one of the edges of the image. The method proposed in~\cite{Hauagge2012} is able to match images of architectural objects taken many years apart and even sketches to photos. The dataset introduced in the paper contains 46 pairs of images.

Matching of images depicting very different appearance of the same object arise in computer vision applications.
A system for guided drawing of free-form objects called Shadow-Draw is presented in \cite{Lee2011}. It can be seen as a large-scale image retrieval system which interactively tries to look for images based on sketches given by a user. In the object classification field, the multiple-appearance problem has been investigated in \cite{Shrivastava2011} who train  a data-driven visual similarity measure in order to match images to sketches or paintings. Those two approaches use global image description rather than local image feature matching.

\vspace{-1.5em}
\section{Datasets}
\label{sec:protocol}
Datasets used in experiments are listed in Table~\ref{tab:all-datasets}. When evaluating detectors (Section~\ref{sec:results}) and the proposed matching algorithm (Section~\ref{sec:wxbs-matcher}) all dataset images are used. However,  descriptor evaluation is performed only on a subset of the most challenging and prominent pairs (i.e. only pairs 1-6 from OxfordAffine) with provided homography of each \textsc{WxBS}category.

Most of the published datasets (with exception of the LostInPast dataset \cite{Fernando2014}) include only a single nuisance factor per image pair.
This is suitable for evaluation of the robustness to a particular nuisance factor but fails to predict performance in more complex environments. One of the motivations of the proposed \text{WxBS} datasets is to address this issue.
\begin{table}[htb]
\caption{Datasets used for evaluation}
\label{tab:all-datasets}
\vspace{1em}
\scriptsize
\centering
\setlength{\tabcolsep}{.3em}
\begin{tabular}{llll}
\hline
Short name& Proposed by& \#images&Type\\
\hline
GDB&Kelman~\etal~\cite{Kelman2007}, 2007&22 pairs& \textsc{WlBS}, \textsc{WsBS}\\
SymB&Hauagge and Snavely~\cite{Hauagge2012}, 2012&46 pairs&\textsc{WaBS}, \textsc{WlBS}\\
MMS& Aguilera~\etal~\cite{Aguilera2012}, 2012&100 pairs&\textsc{WsBS}\\
EVD&Mishkin~\etal~\cite{Mishkin2013}, 2013&15 pairs&\textsc{WgBS}\\
OxAff&Mikolajczyk~\etal\cite{Mikolajczyk2005},~\cite{Mikolajczyk2005a}, 2013&8 sixplets&\textsc{WgBS}\\
EF&Zitnick and Ramnath~\etal\cite{Zitnick2011},2011&8 sixplets&\textsc{WgBS},\textsc{WlBS}\\
Amos&Jacobs~\etal\cite{Jacobs2007},2007& $>$ 100K&\textsc{WlBS},\textsc{WaBS}\\
VPRiCE&VPRICE Challenge 2015~\cite{VPRICE2015}& 3K pairs&\textsc{WgaBS}, \textsc{WglBS},\textsc{WgsBS},\\
Past&Fernando~\etal\cite{Fernando2014}, 2014& 502 images&\textsc{WgaBS}\\
WxBS&here& 37 pairs& \textsc{WaBS},\textsc{WgaBS},\textsc{WglBS}, \textsc{WgsBS},\textsc{WlaBS},\textsc{WgalBS} \\
\end{tabular}
\vspace{-1em}
\end{table}

\begin{figure*}[t]
\begin{center}
\def\svgwidth{1\textwidth}
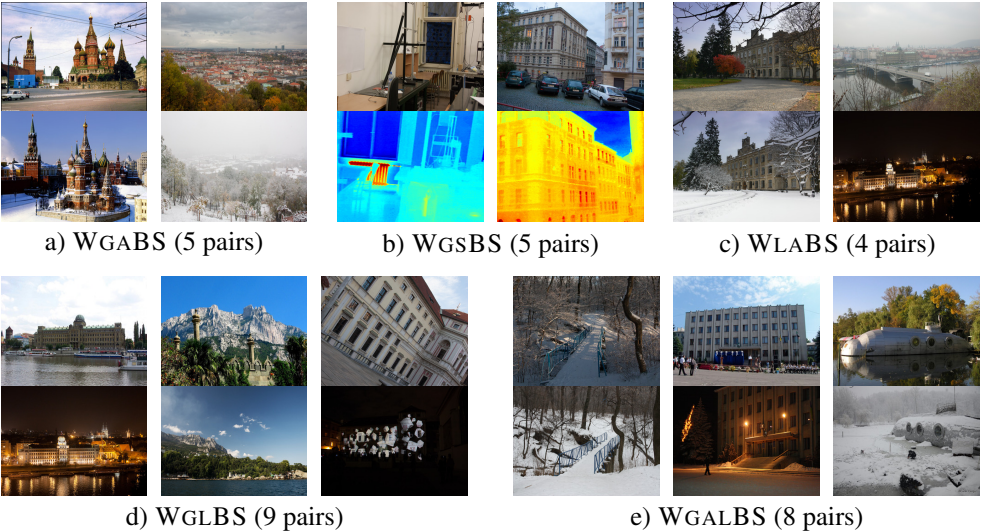
\end{center}
   \caption{Examples of image pairs from the \textsc{WxBS} dataset.}
\label{fig:wxbs-examples}
\vspace{-1em}
\end{figure*}

\noindent
{\bf WxBS dataset and evaluation protocol.} A set of 37 image pairs has been collected from Flickr and other sources. The dataset is divided into 6 categories based on the combinations of nuisance factor present, see~Table~\ref{tab:wxbs-categories}. For every image, a set of approximately 20 ground-truth correspondences has been annotated. Selected examples are presented in Figure \ref{fig:wxbs-examples}.
The resolution of the majority of the images is $800 \times 600$ with the exception of LWIR images from the \textsc{WgsBS} dataset which were captured by a thermal camera with a resolution of $250 \times 250$ pixels.
The selected image pairs contain both urban and natural scenes.
\begin{table}[htb]
\caption{The WxBS datasets categories}
\vspace{0.5em}
\label{tab:wxbs-categories}
\footnotesize
\centering
\setlength{\tabcolsep}{.3em}
\begin{tabular}{llrr}
\hline
Short name & Nuisance & \#images & Avg. \# GT Corr. \\
\hline
\textsc{map2ph} & appearance (map to photo) & 6 pairs & homography provided\\
\textsc{WgaBS} & viewpoint, appearance & 5 pairs & 22 per img.\\
\textsc{WglBS} & viewpoint, lighting & 9 pairs & 21 per img.\\
\textsc{WgsBS} & viewpoint, modality & 5 pairs & 18 per img.\\
\textsc{WlaBS} & lighting, appearance& 4 pairs & 25 per img.\\
\textsc{WgalBS} & viewpoint, appearance, lighting& 8 pairs & 17 per img.\\
\end{tabular}
\end{table}

\noindent
{\bf Ground truth and the evaluation protocol.}
In the image registration tasks, it is often sufficient to define ground truth as a homography between an image pair. However, the WxBS dataset contains significant viewpoint changes. In the case of a non-planar scene  a homography can, at best, cover the dominant plane. 

We assume that an ideal algorithm matches the majority of the scene content, thus our ground truth is a set of manually selected correspondences which evenly cover the part of the scene visible in both images. The average number of correspondences per image pair is shown in Table \ref{tab:wxbs-categories}. 

\paragraph{The evaluation protocol for the WxBS dataset.}
For each image pair indexed with  $i \in \mathbb{Z}$  we have manually annotated a set of correspondences $(\mathbf{u}_i,\mathbf{v}_i) \in C_i$ where $\mathbf{u}$ and $\mathbf{v}$ are positions in the 1\textsuperscript{st} and the 2\textsuperscript{nd} image respectively. 
For epipolar geometry we use the \emph{symmetric epipolar distance}  and the \emph{symmetric reprojection error} for homography \cite{Hartley2000}.

Recall on ground truth correspondences $C_i$ of image pair $i$ and for geometry model $\mathbf{M}_i$ is computed as a function of a threshold~$\theta$
\begin{equation}
 \mathrm{r}_{i,\mathbf{M}_i}(\theta) = \frac{| (\mathbf{u}_i,\mathbf{v}_i) : (\mathbf{u}_i,\mathbf{v}_i) \in C_i,  e(\mathbf{M}_i,\mathbf{u},\mathbf{v}) < \theta|}{| C_i |}
\end{equation}
using appropriate error functions. For all pairs of each category $W$ we define an overall recall per category as:
\begin{equation}
 \mathrm{r}_{W}(\theta) = \frac{1}{| W | \sum_{i \in W} \mathrm{r}_{i,M_i}(\theta)}
\end{equation}
This measure is as the fraction of the confirmed annotated correspondences for a given threshold in a nuisance category. 

\begin{wrapfigure}{L}{0.5\textwidth}
\vspace{-2em}
\begin{spacing}{0.2}
\begin{minipage}{0.5\textwidth}
\begin{algorithm}[H]
\caption{\textsc{MODS-WxBS} -- a matcher for wide multiple baseline stereo}
\label{alg:wxbs-m}
\begin{algorithmic}
\footnotesize
\Require{$I_1$, $I_2$ -- two images; $\theta_m$ -- minimum required number of matches; $S_{\text{max}}$ -- maximum number of iterations.  }
\Ensure{Fundamental or homography matrix F or H;\\
 a list of corresponding local features.}
\algrule
\While {$(N_{\text{matches}} < \theta_{m})$ ${\mathit and}$ $(\text{Iter} < S_{\text{max}})$} 
\For {$I_1$ and $I_2$ separately}
\State \textbf{1 Generate synthetic views} according to the\\
\hskip3em scale-tilt-rotation-detector setup for the Iter.
\State \textbf{2 Detect local features} using adaptive threshold.
\State \textbf{3} Extract rotation invariant descriptors with: \\
\hskip3em \textbf{3a rSIFT} \hskip1em and \hskip1em \textbf{3b hrSIFT}
\State \textbf{4 Reproject local features} to $I_1$.
\EndFor
\State \textbf{5 Generate tent. corresp.} based on the first geom. \\ 		    \hskip3em inconsistent rule for rSIFT and hrSIFT \\ 
    \hskip3em separately using kD-tree
\State \textbf{6 Filter duplicates}
\State \textbf{7 Geometric verification} of \emph{all} TC with modified \\ 
    \hskip3em DEGENSAC estimating $F$ or $H$.
\State \textbf{8 Check geom. consistency} of the LAFs \\ with est. $F$.
\EndWhile
\end{algorithmic}
\end{algorithm}
\end{minipage}
\end{spacing}
\end{wrapfigure}
\section{Matching algorithm for wide multiple baseline stereo}
\label{sec:wxbs-matcher}
\renewcommand\algorithmicindent{1em}
In this section, we propose a variant of MODS~\cite{Mishkin2013, Mishkin2015} matcher designed for WxBS problems called WxBS-MODS, or WxBS-M in short.  Its overall structure is shown in Algorithm~\ref{alg:wxbs-m}. The view synthesis is identical to the original MODS framework \cite{Mishkin2013}.

Tentative correspondences are generated using kD-tree~\cite{Muja2014} and the 1st geometrically inconsistent rule with radius equal 10 pixels as threshold is applied\cite{Mishkin2013}. Descriptors from different detectors types (Hessian, MSER+, MSER-) as well as for different descriptors are put in seperate kD-trees. 
After matching, all tentative correspondences are put into a single list and duplicates, which appears due to view synthesis, are filtered if features in both images are within a 3 pixel radius.

\section{Evaluation of description and detection algorithms}
\label{sec:results}
In this section, multiple detection and description algorithms are evaluated. 

\noindent
{\bf Descriptors evaluation.}
The evaluation protocol is as follows. The dataset consists of 40 image pairs from datasets listed in Table~\ref{tab:all-datasets} divided into 5 parts
by the nuisance factor. For all pairs, homography is the appropriate two-view relationship -- the images are either without significant relative depth of taken from virtually identical viewpoints.  In order to minimize bias towards a specific detector, affine-covariant regions by Hessian-Affine, MSER and FOCI in the first -- least challenging image of the pair are used (visible in case of IR-vis, day on day-night, frontal when view point changes, etc.). 
The affine-covariant regions have been detected with dominant orientation and then reprojected to the second image by the ground truth homography. Features which are not visible in the second image have been discarded. Therefore geometric repeatability of affine regions on the selected regions is always $100\%$ and the maximum possible recall is 1. Color-to-grayscale image transformation have been done via channel averaging, which gives best matching performance~\cite{Kanan2012}.

Then affine regions were normalized to patch size 41x41 (scale $\sigma = 3\sqrt{3}$) and described with given descriptors. An affine-normalization procedure is performed even for the fast binary descriptors, which is rarely used because of the significant additional processing time. However, the goal of our experiment is to explore descriptor performance in challenging conditions, not their speed. The procedure helps -- the typical threshold of the Hamming distance for binary descriptors  on unnormalized patch is around 60-80, while on affine normalized patches similar performance is obtained with a threshold around 10-30. All descriptors clearly benefit from the affine-normalized process, e.g. the graffiti 1-6 pair from the OxfordAffine dataset could be matched with FREAK descriptor only when using a normalized patch.

The tested descriptors are: SIFT~\cite{Lowe2004}, rSIFT~\cite{Arandjelovic2012}, hrSIFT (gradients in interval $\left[0;\pi\right)$)~\cite{Kelman2007}, InvSIFT (SIFT with reordered cells as for inverted image)~\cite{Hare2011}, LIOP\cite{ZhenhuaWang2011}, AKAZE~\cite{Alcantarilla2013}, MROGH~\cite{Fan2012}, FREAK~\cite{Alahi2012}, ORB~\cite{Rublee2011}, SymFeat~\cite{Hauagge2012}, SSIM~\cite{Shechtman2007} (implementation \cite{Chatfield2009}), DAISY~\cite{Tola2010} and $L_2$-normalized raw grayscale pixel intensities. 
\newcommand{\smallImgWidth}{0.16}
\begin{figure*}[tb]
\centering
\subfigure[\texttt{WGBS}]{\includegraphics[width=0.16\textwidth]{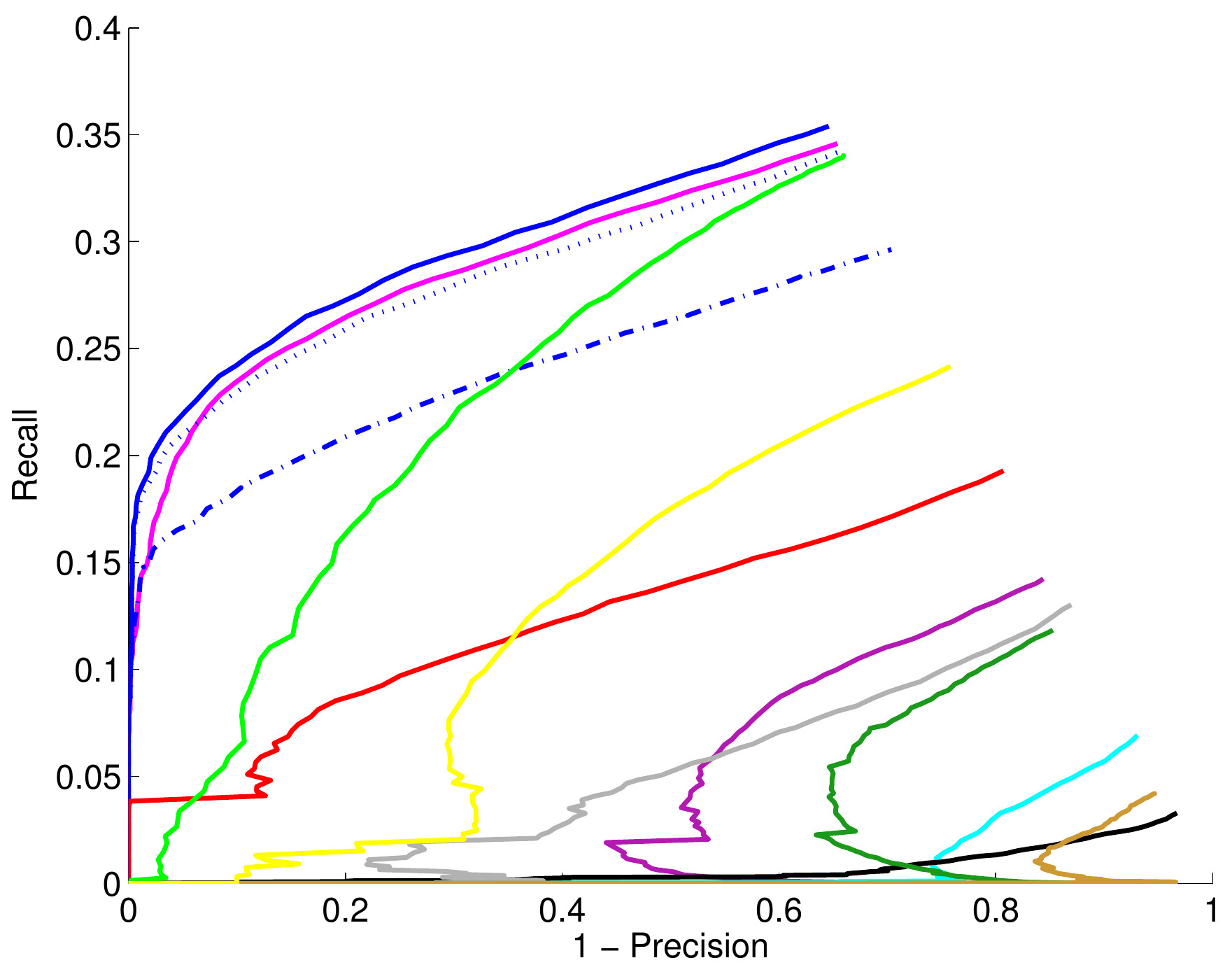}}
\subfigure[\texttt{WLBS}]{\includegraphics[width=0.16\textwidth]{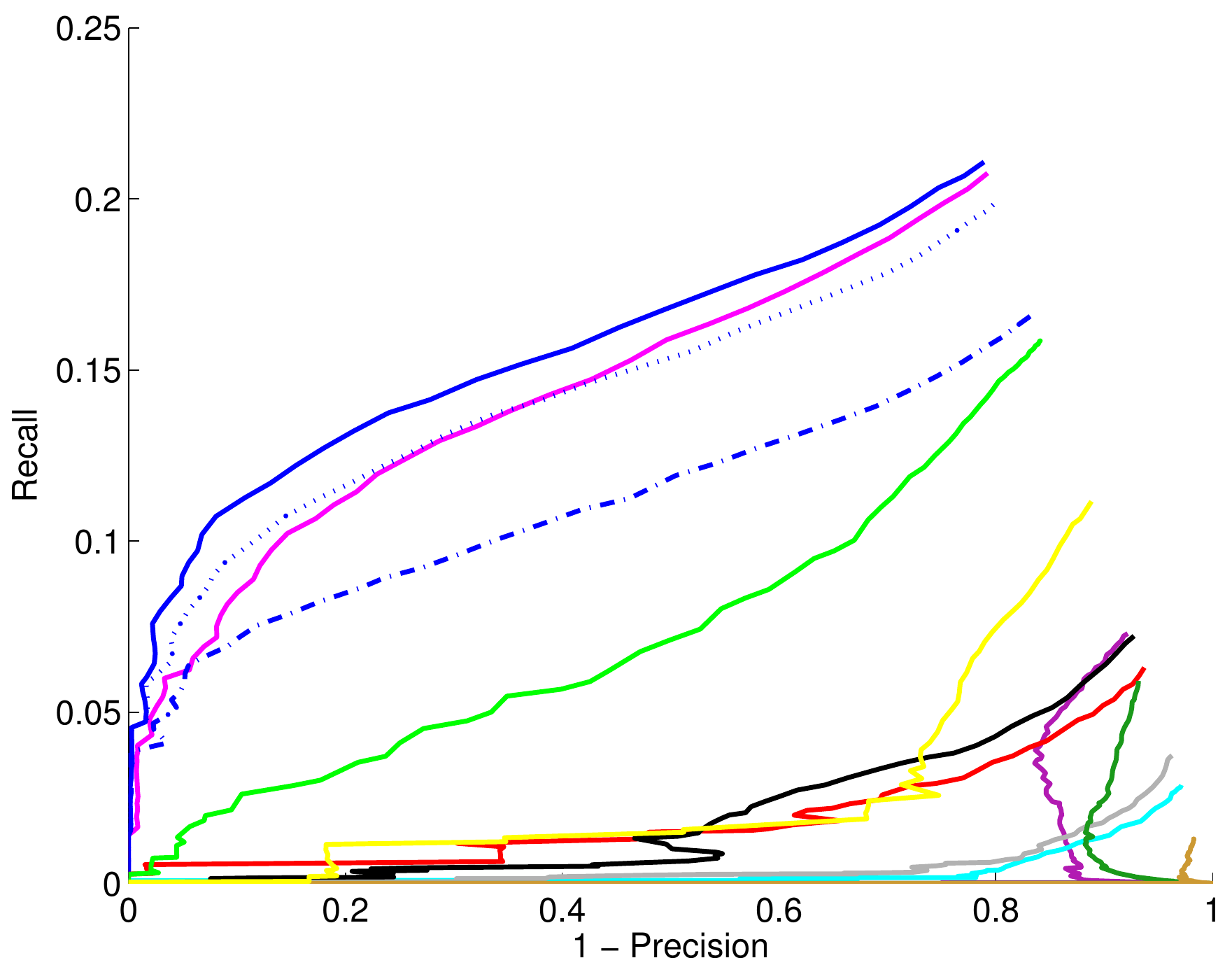}}
\subfigure[\texttt{WLBS}]{\includegraphics[width=0.16\textwidth]{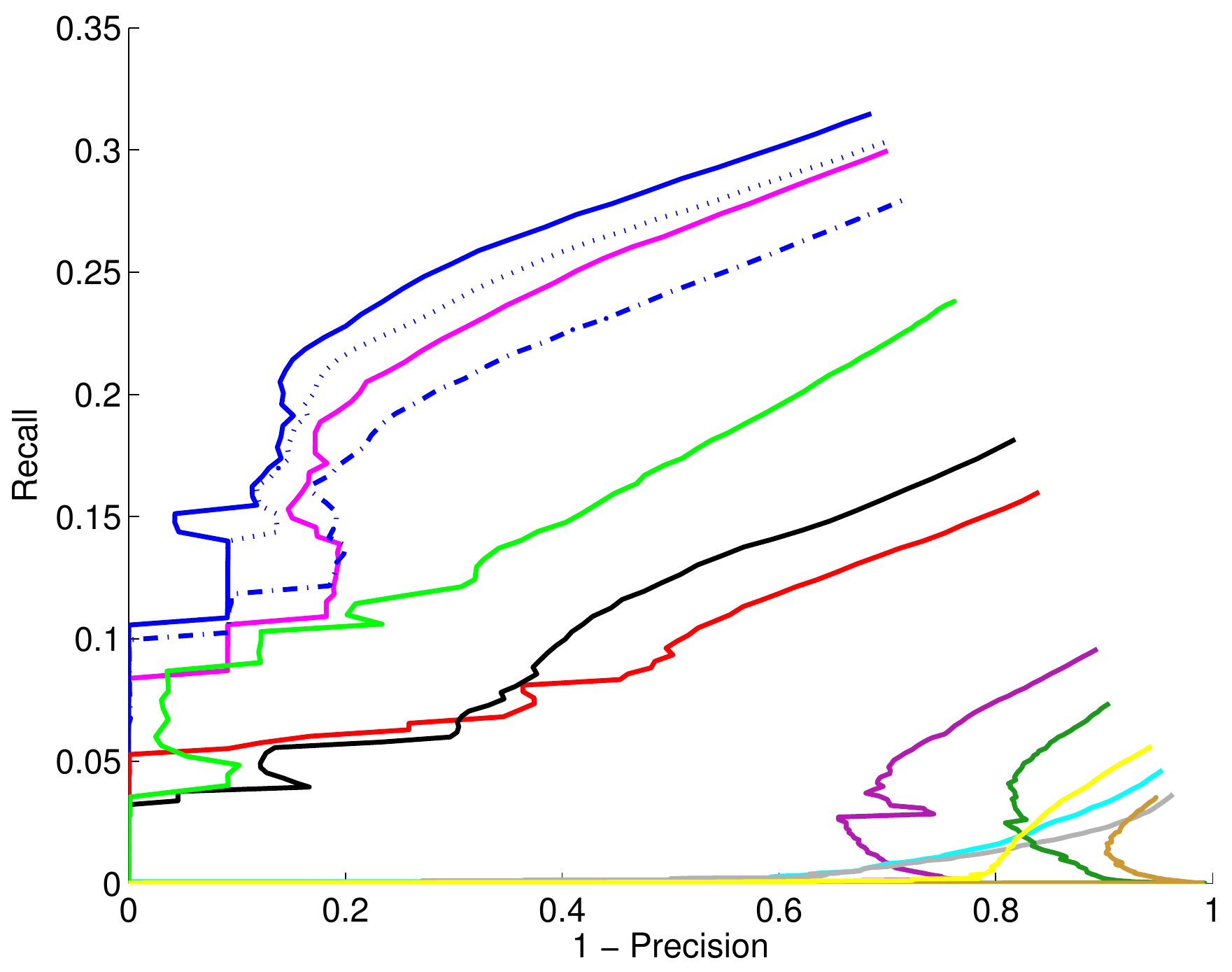}}
\subfigure[\texttt{WSBS}]{\includegraphics[width=0.16\textwidth]{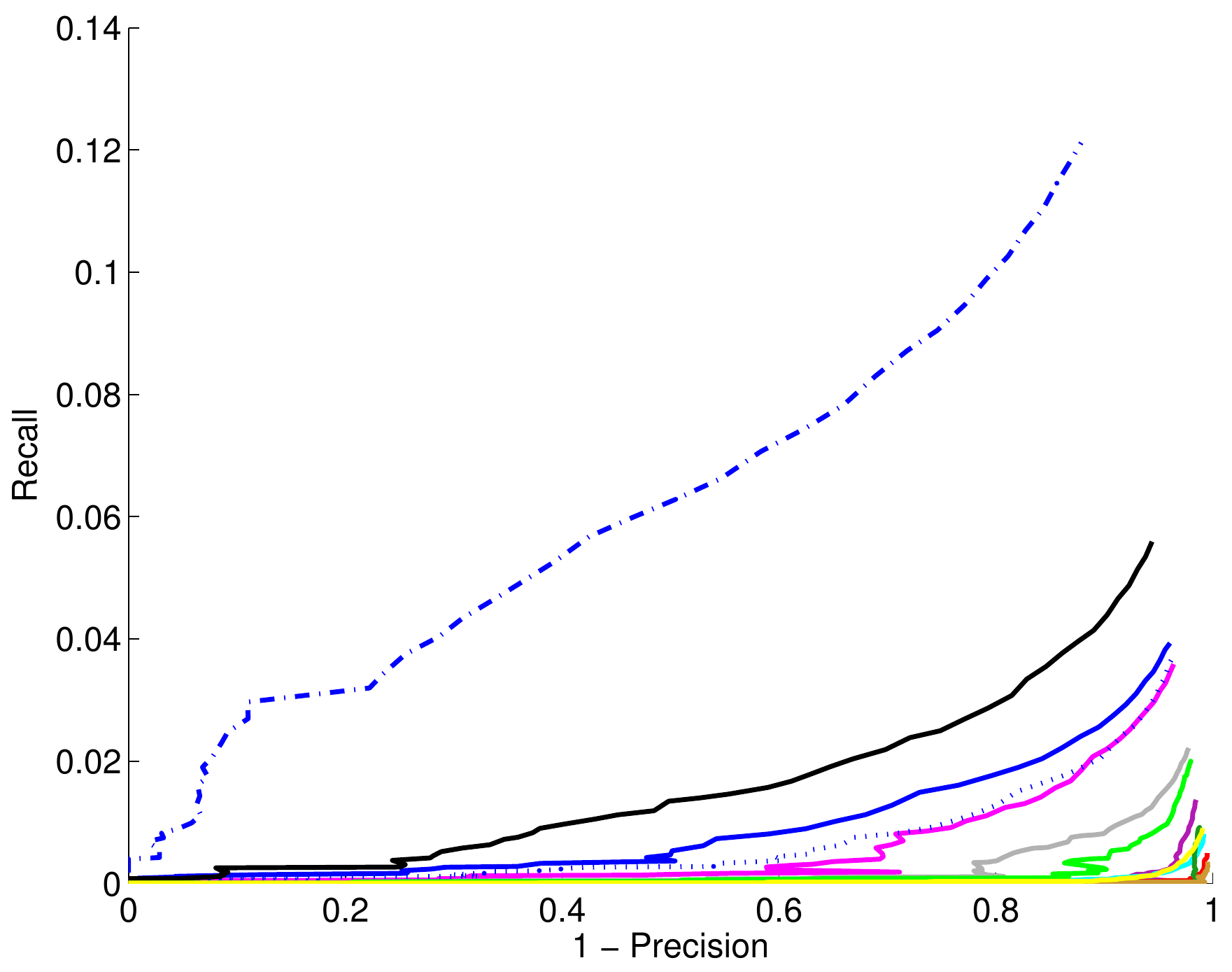}}
\subfigure[\texttt{map2ph}]{\includegraphics[width=0.16\textwidth]{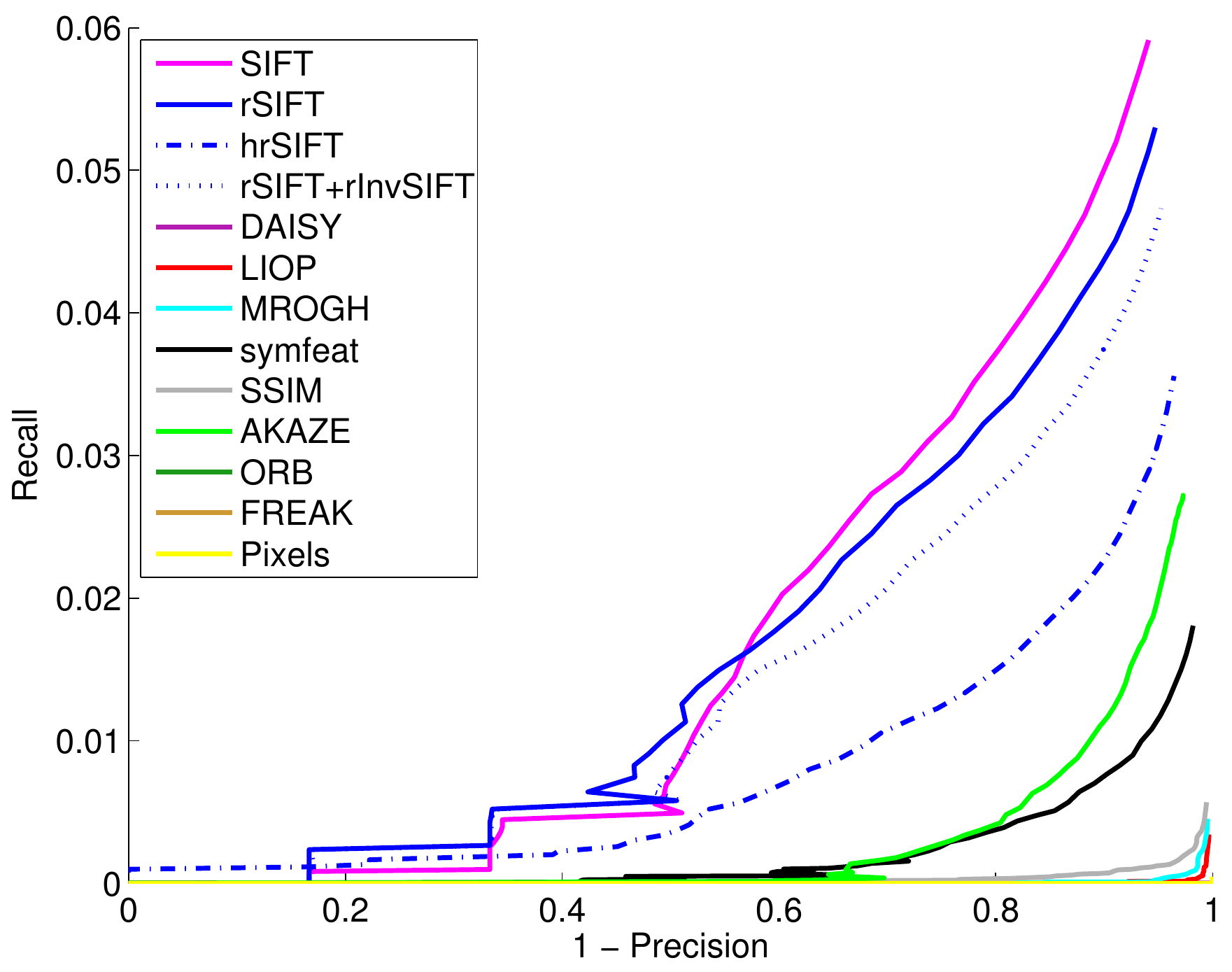}}
\subfigure[\texttt{all}]{\includegraphics[width=0.16\textwidth]{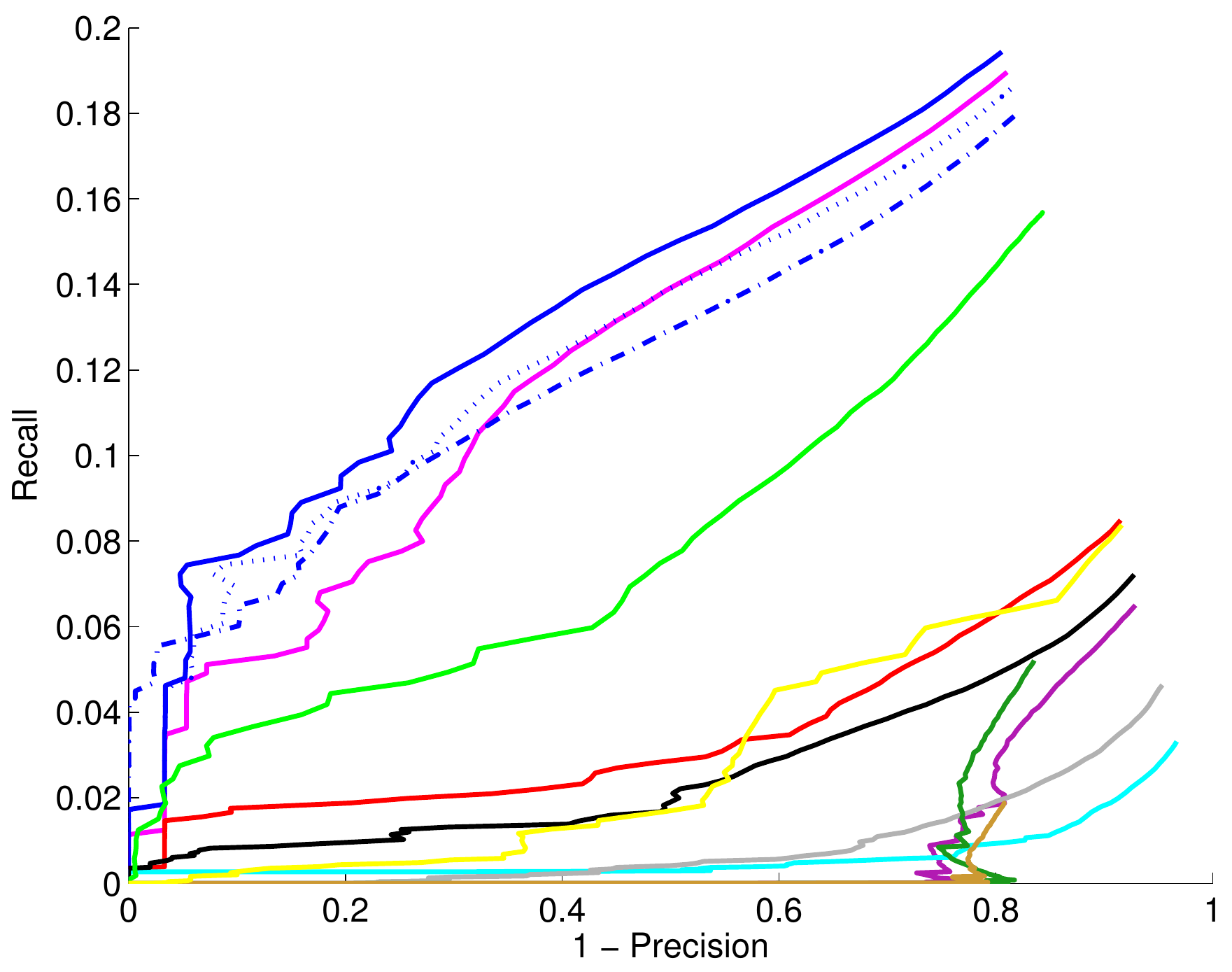}}
\\
\subfigure[\texttt{WGBS}]{\includegraphics[width=0.16\textwidth]{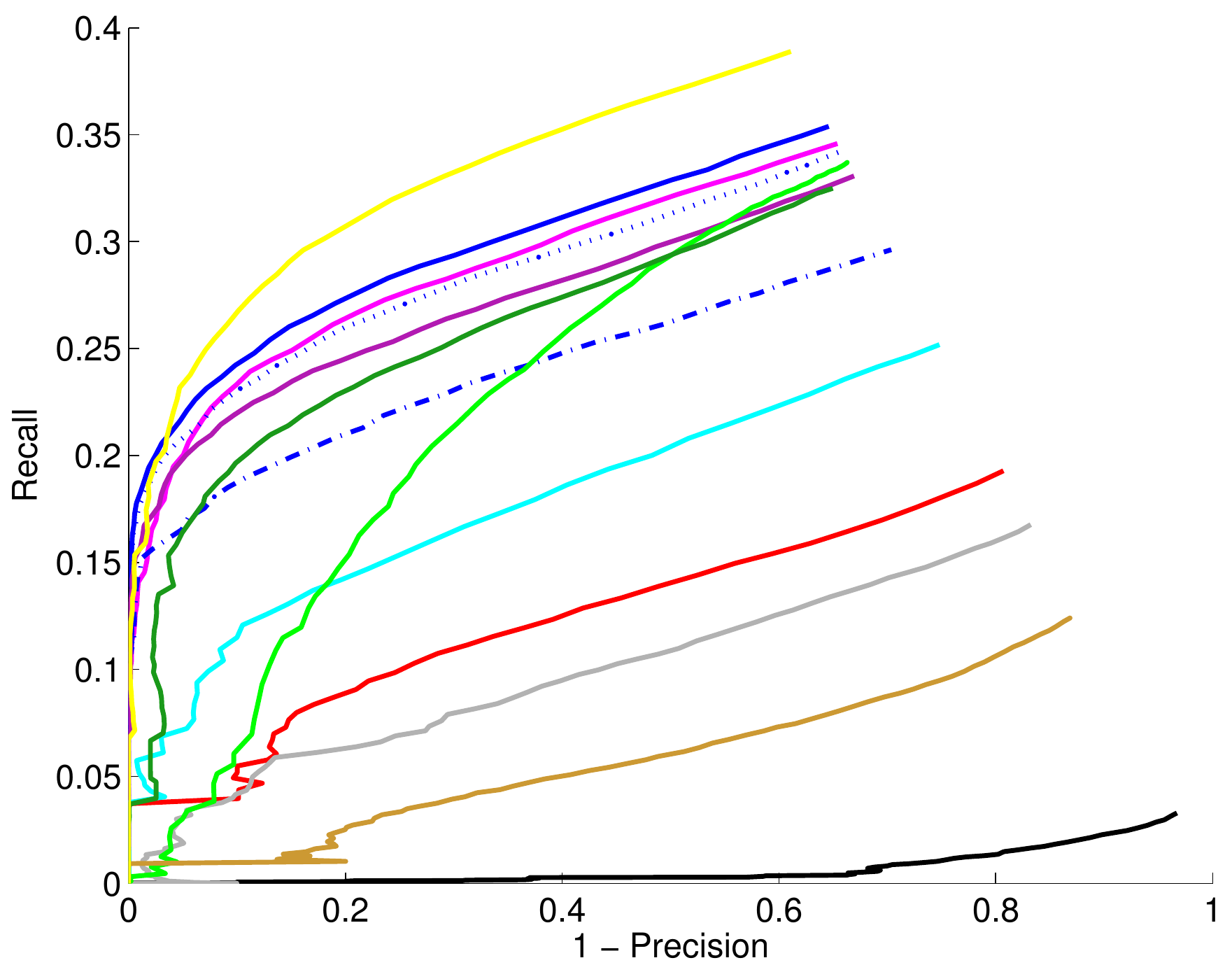}}
\subfigure[\texttt{WLBS}]{\includegraphics[width=0.16\textwidth]{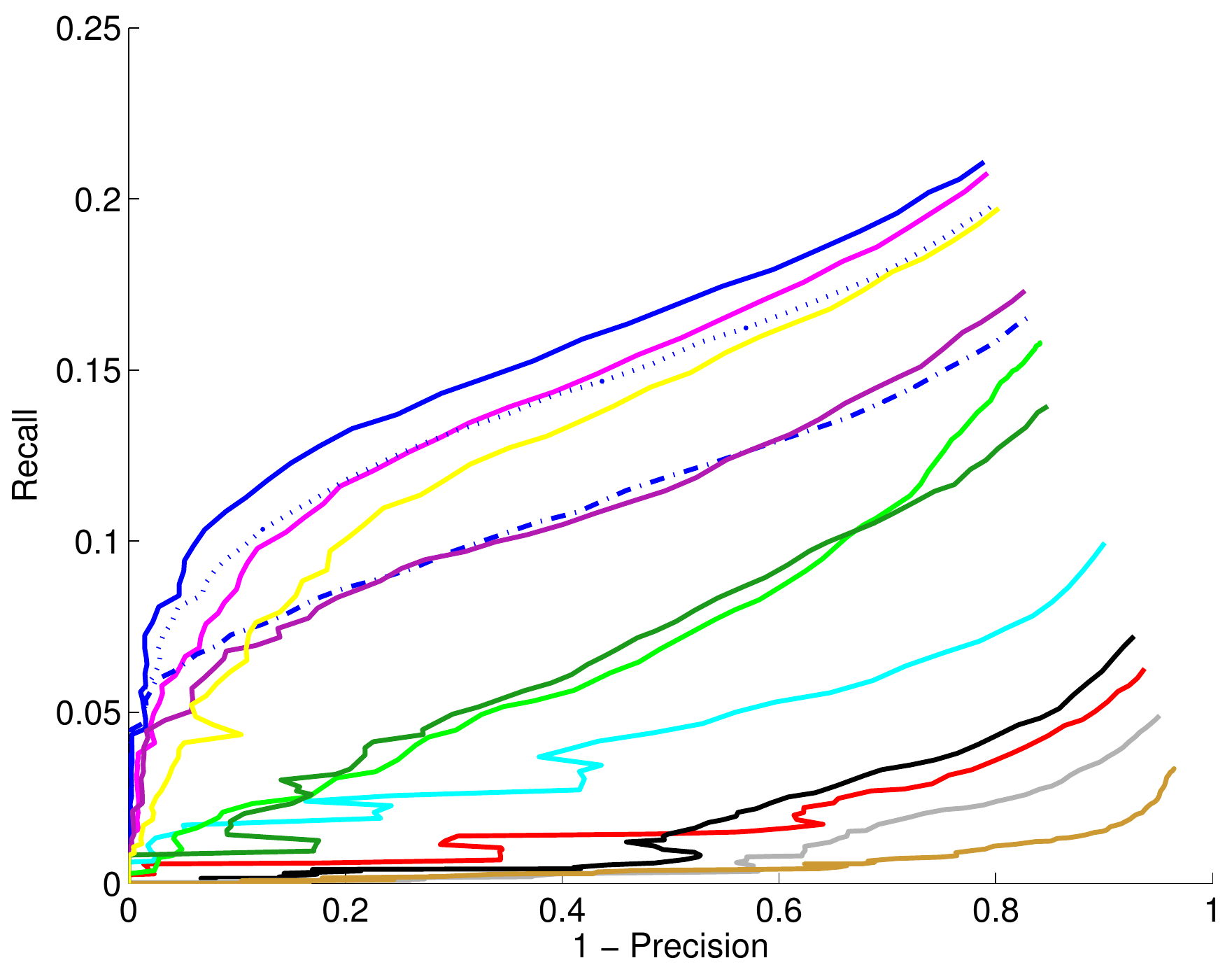}}
\subfigure[\texttt{WLBS}]{\includegraphics[width=0.16\textwidth]{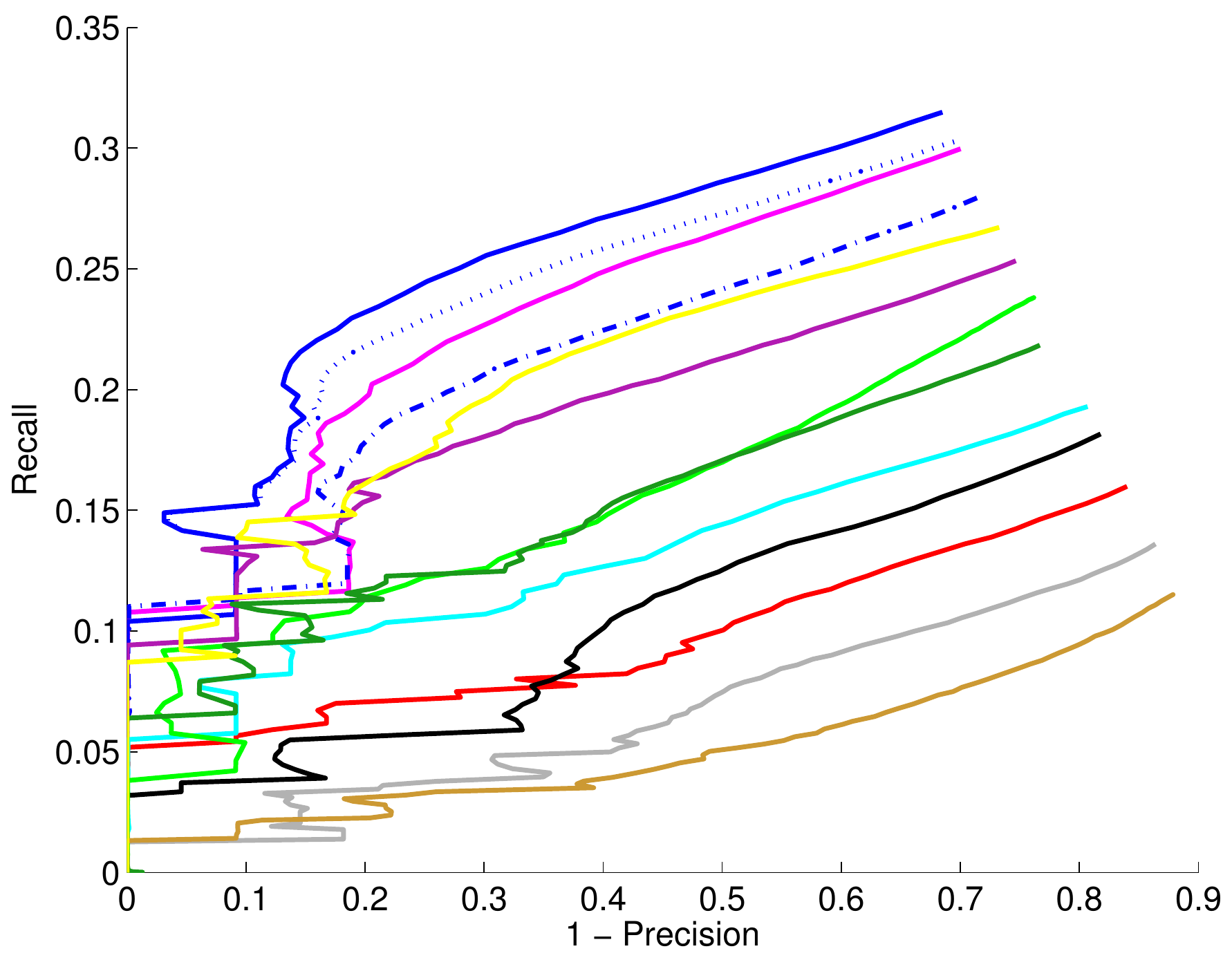}}
\subfigure[\texttt{WSBS}]{\includegraphics[width=0.16\textwidth]{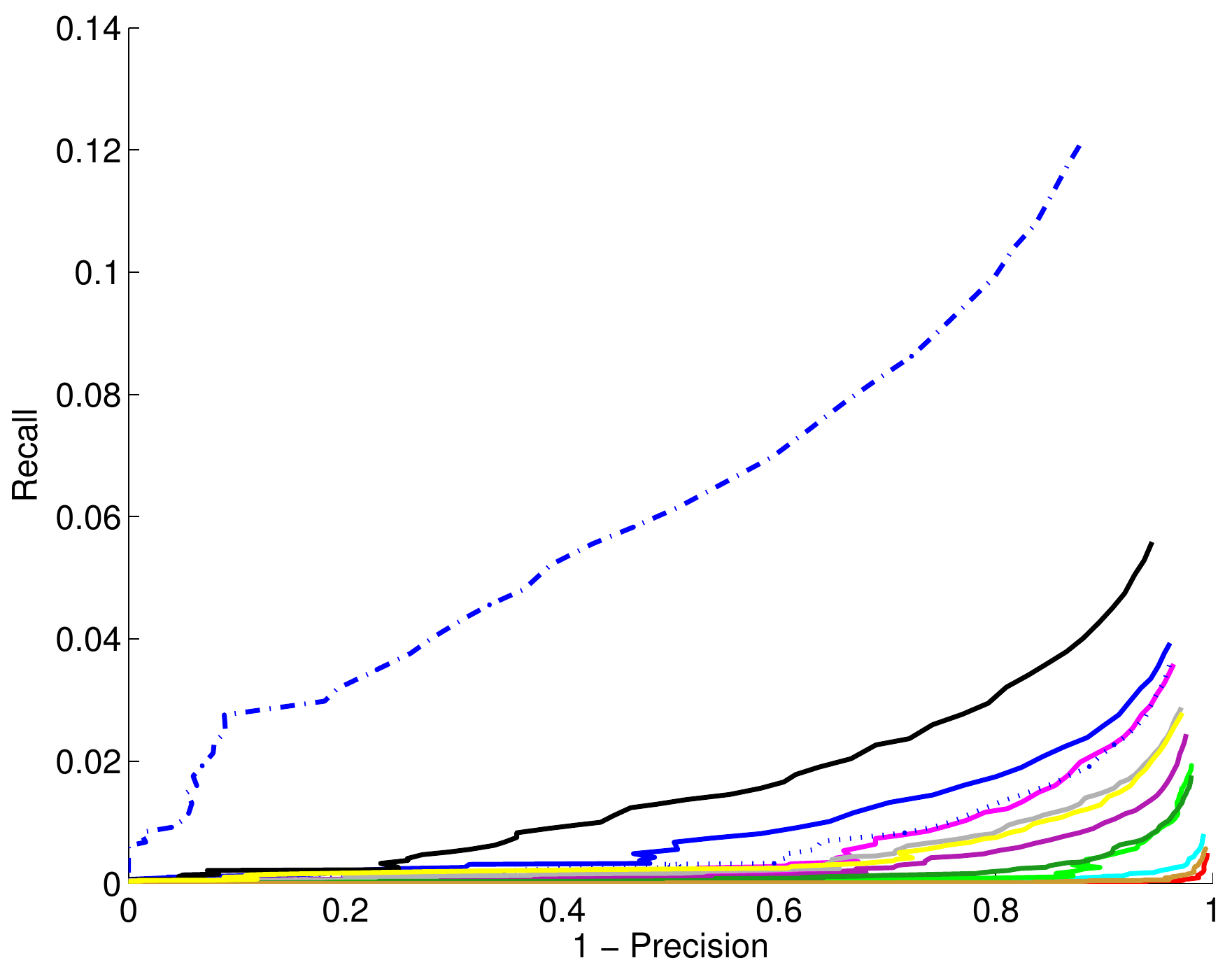}}
\subfigure[\texttt{map2ph}]{\includegraphics[width=0.16\textwidth]{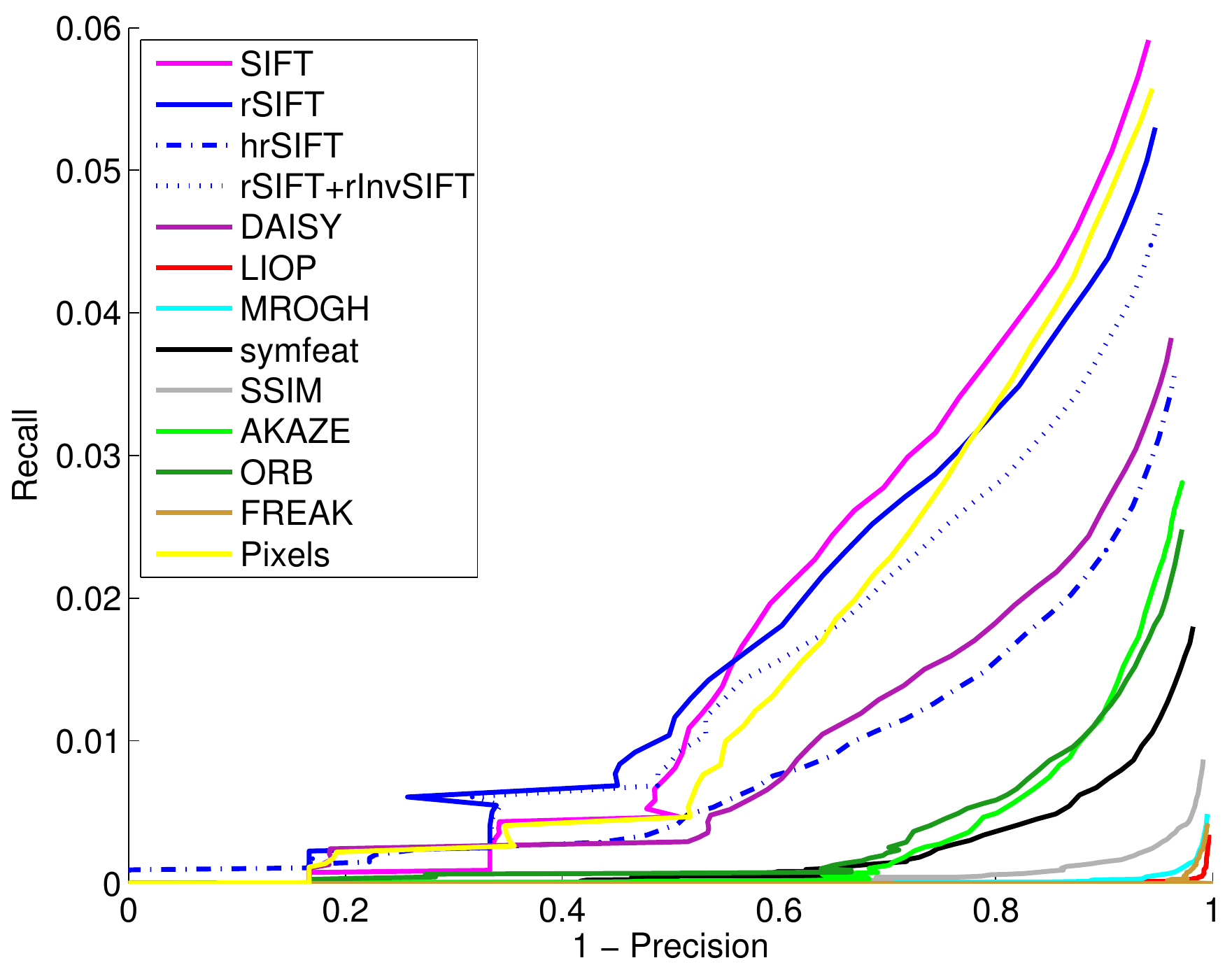}}
\subfigure[\texttt{all}]{\includegraphics[width=0.16\textwidth]{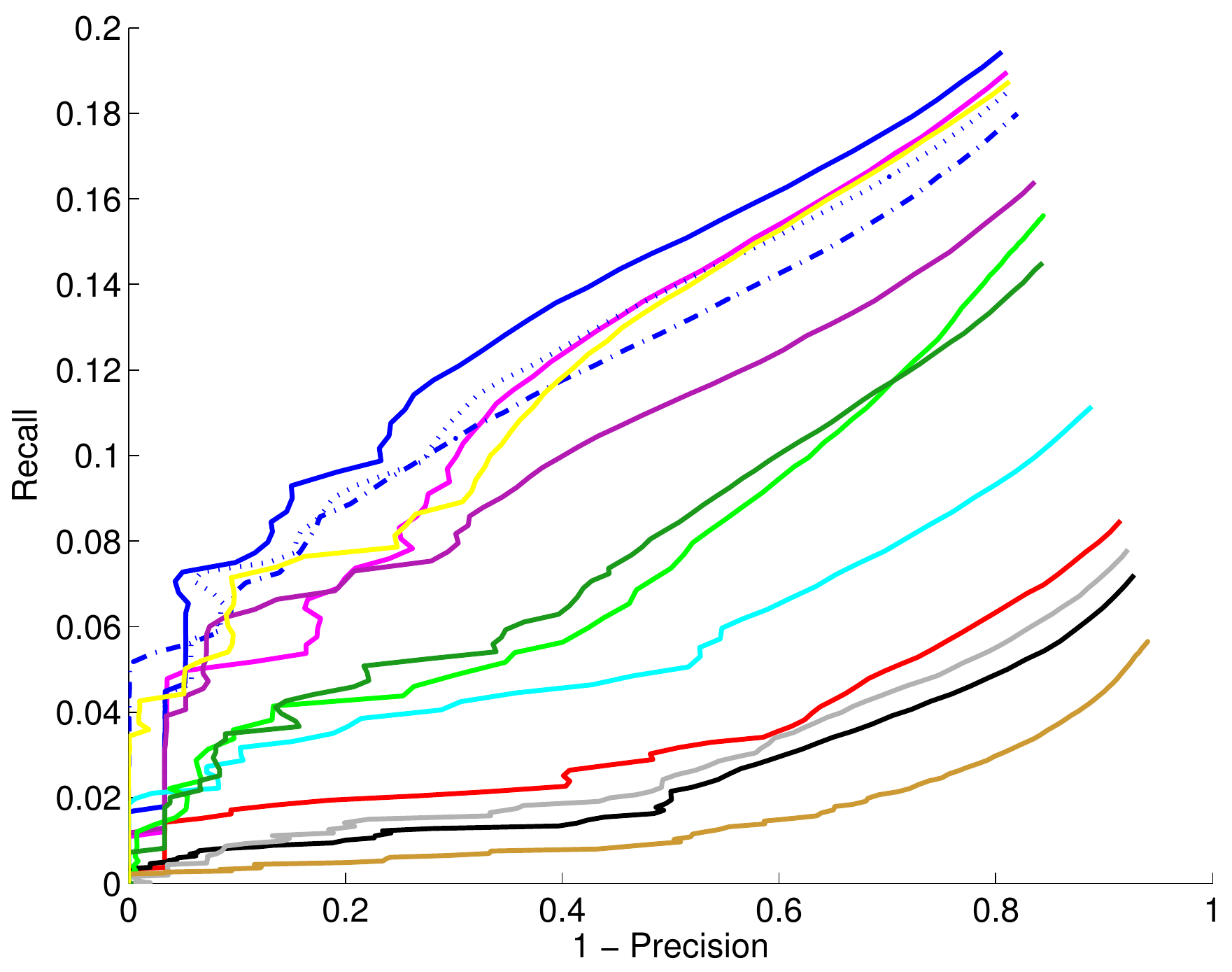}}
\\
\subfigure[\texttt{WGBS}]{\includegraphics[width=0.16\textwidth]{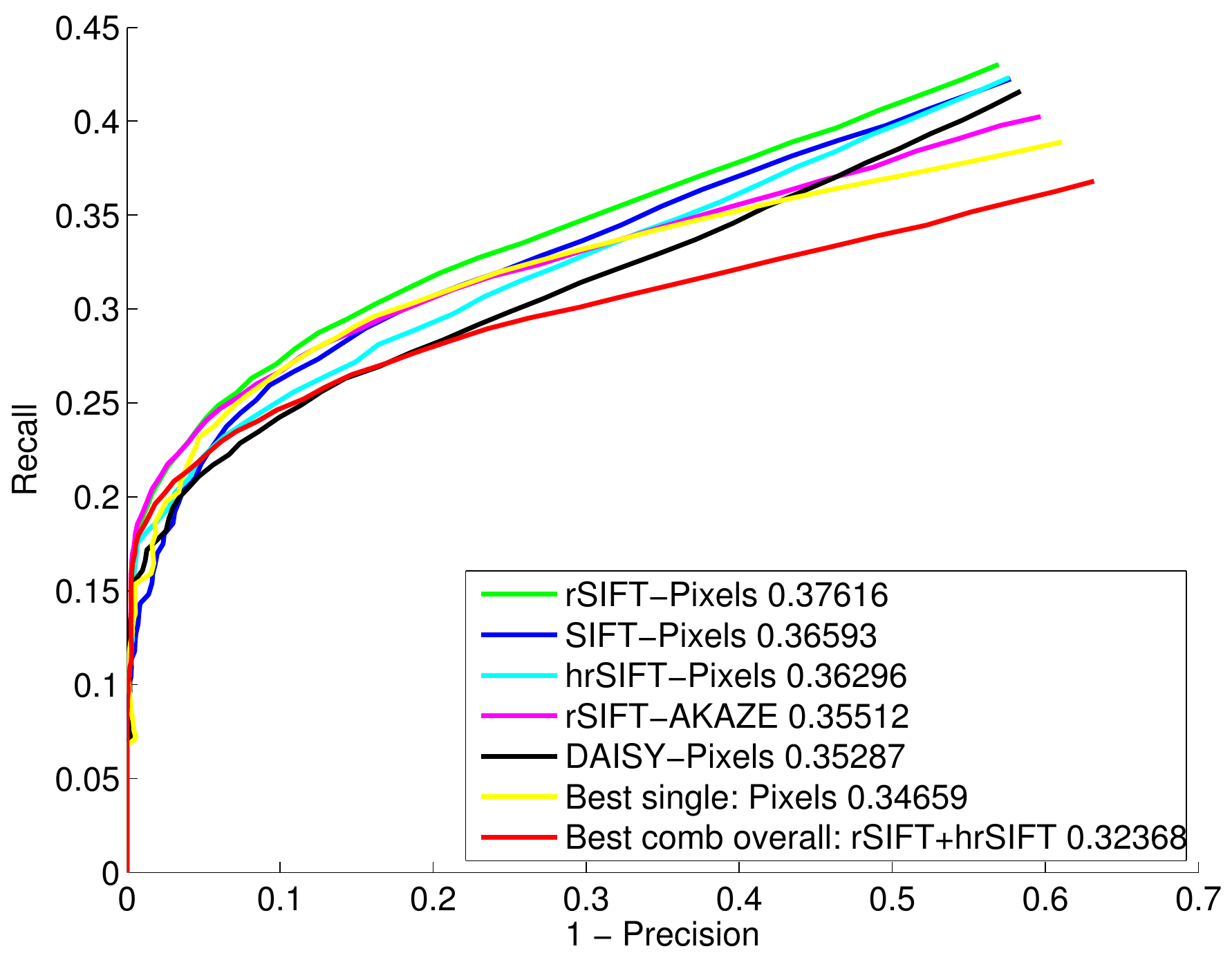}}
\subfigure[\texttt{WLBS}]{\includegraphics[width=0.16\textwidth]{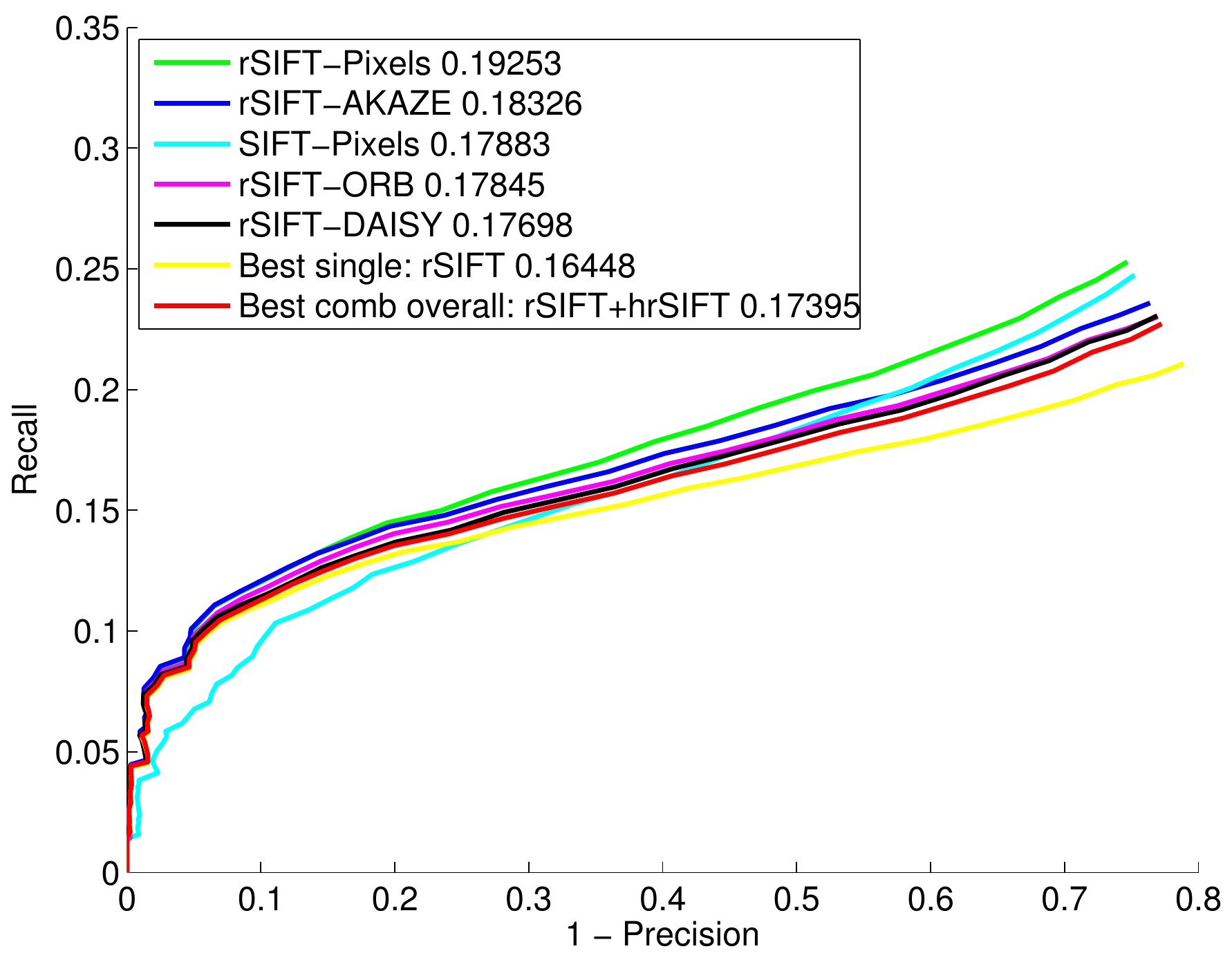}}
\subfigure[\texttt{WLBS}]{\includegraphics[width=0.16\textwidth]{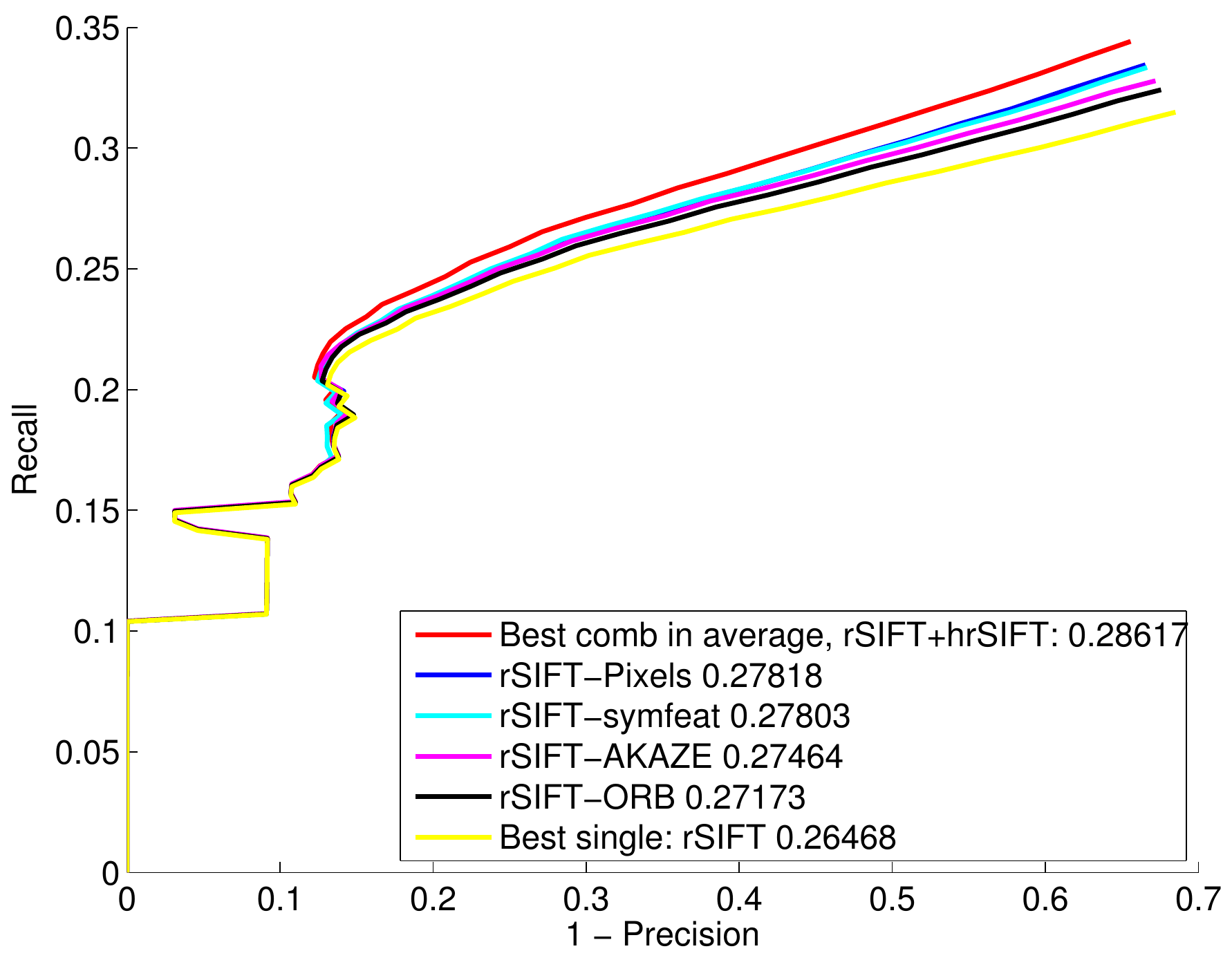}}
\subfigure[\texttt{WSBS}]{\includegraphics[width=0.16\textwidth]{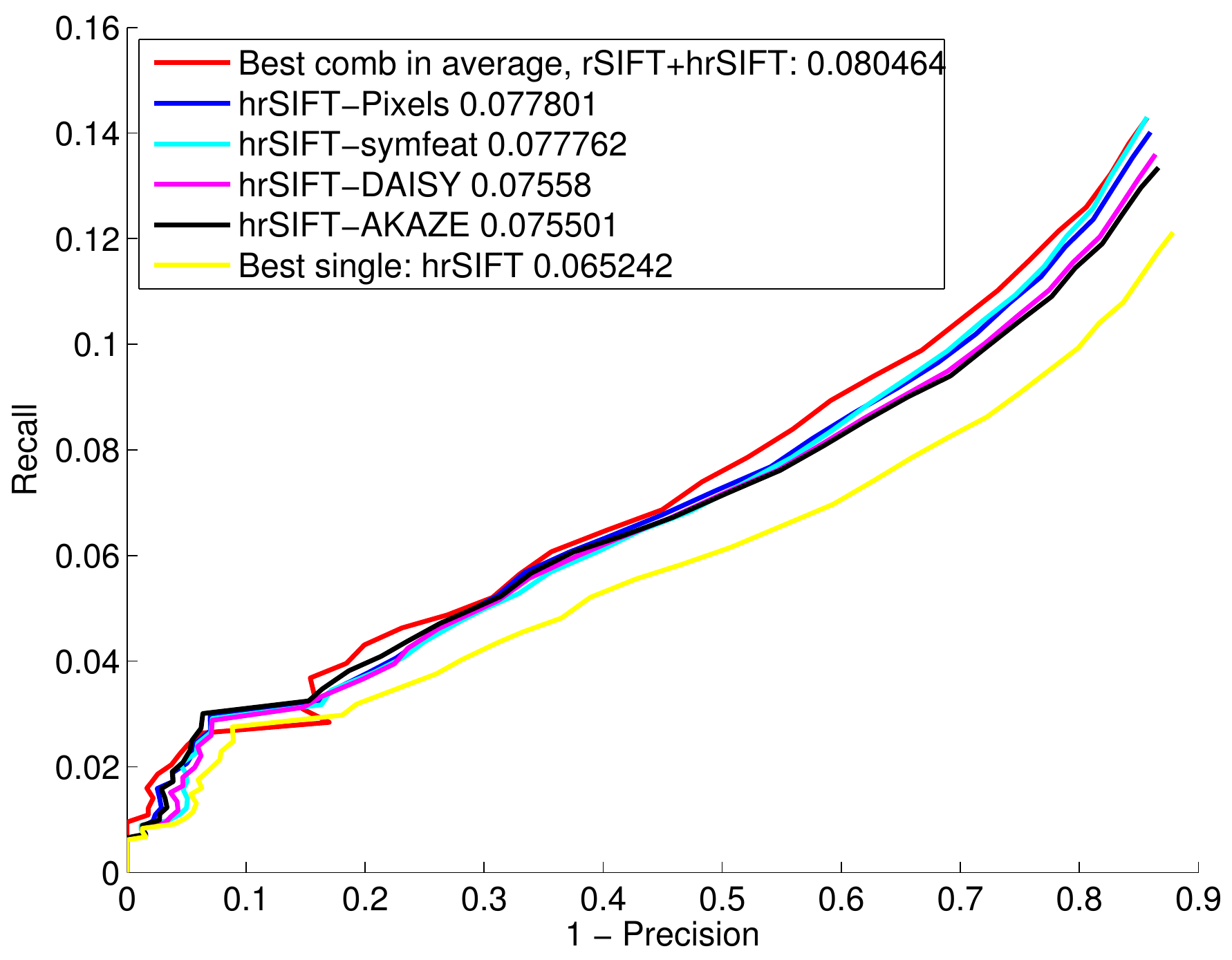}}
\subfigure[\texttt{map2ph}]{\includegraphics[width=0.16\textwidth]{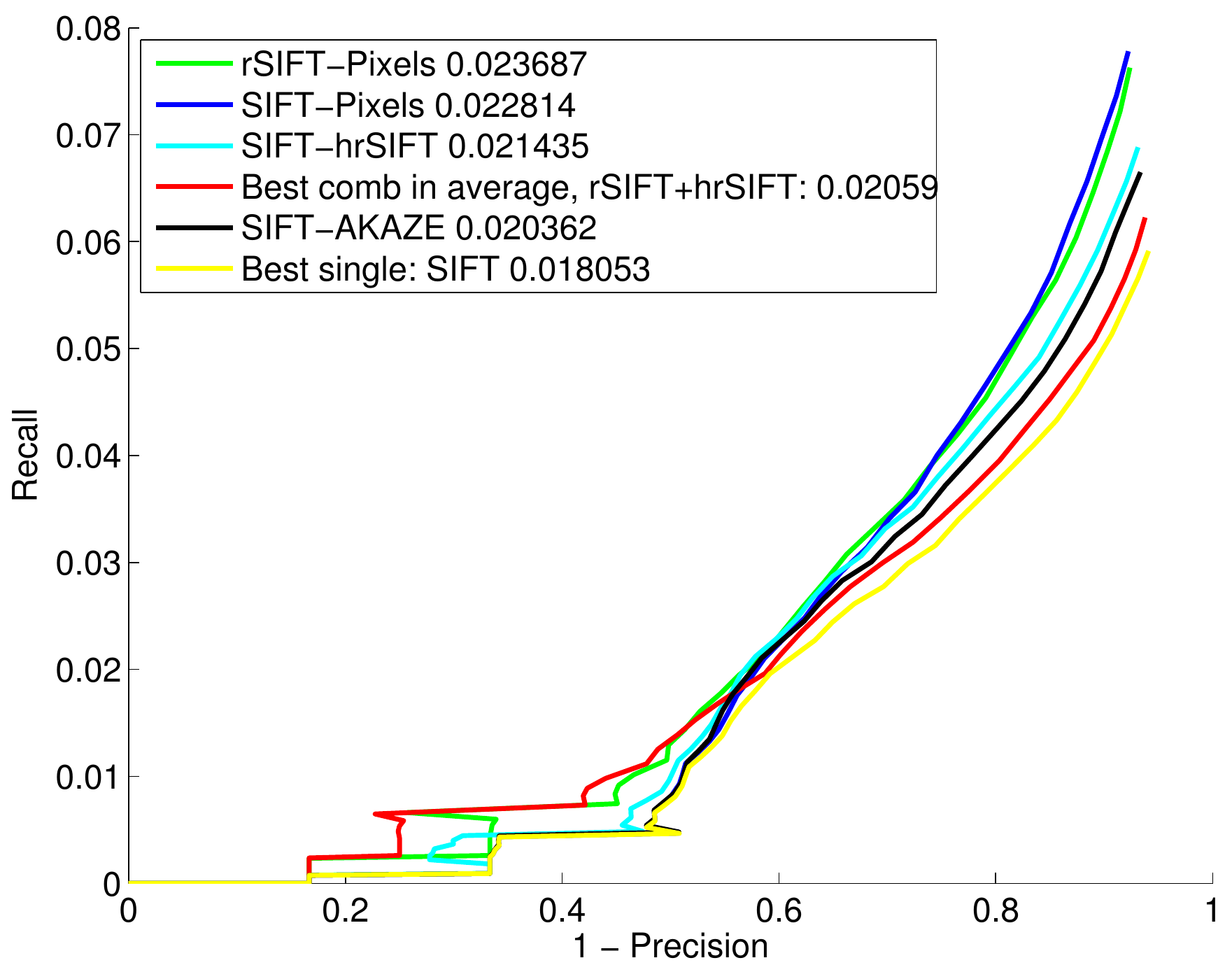}}
\subfigure[\texttt{all}]{\includegraphics[width=0.16\textwidth]{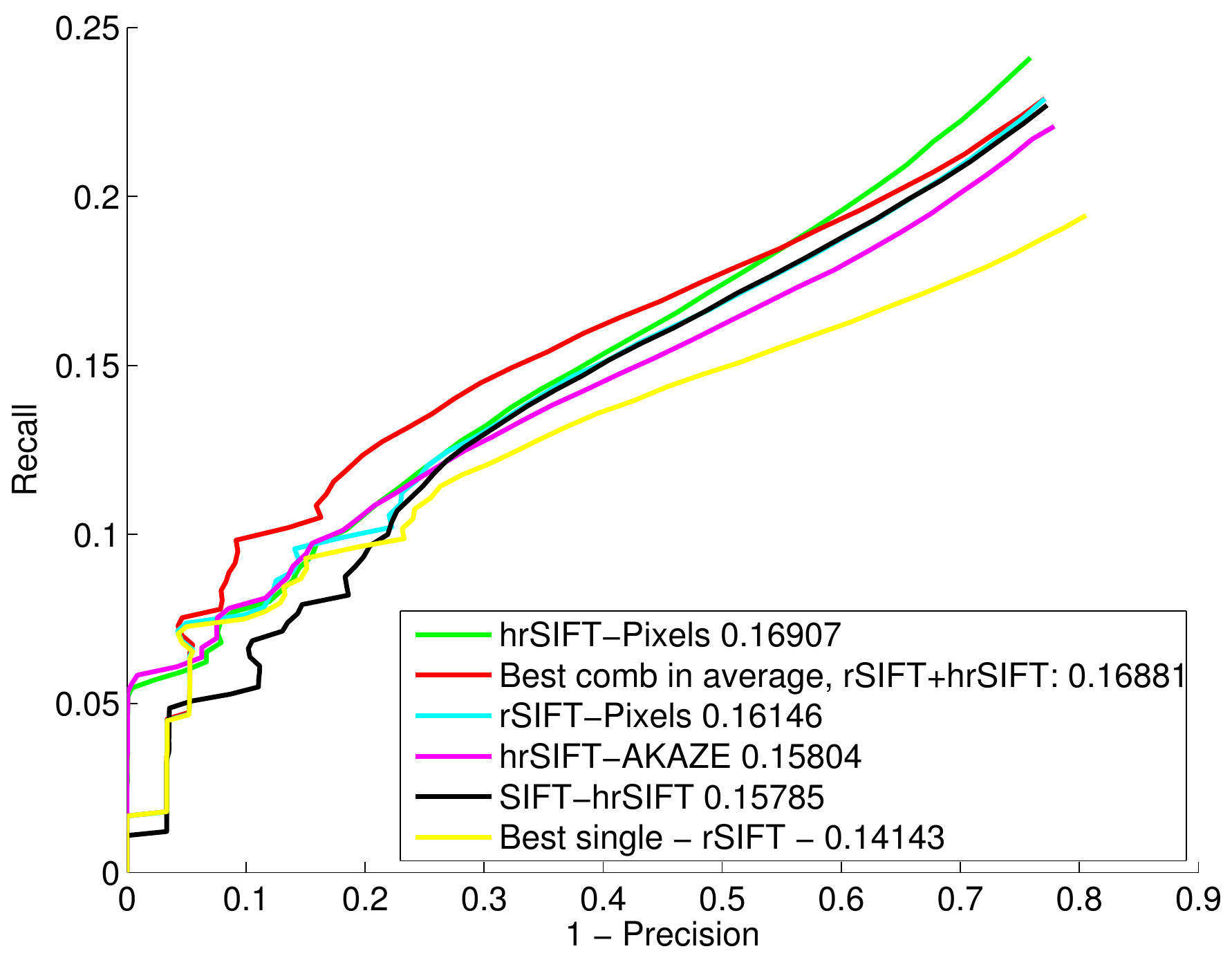}}
\begin{center}
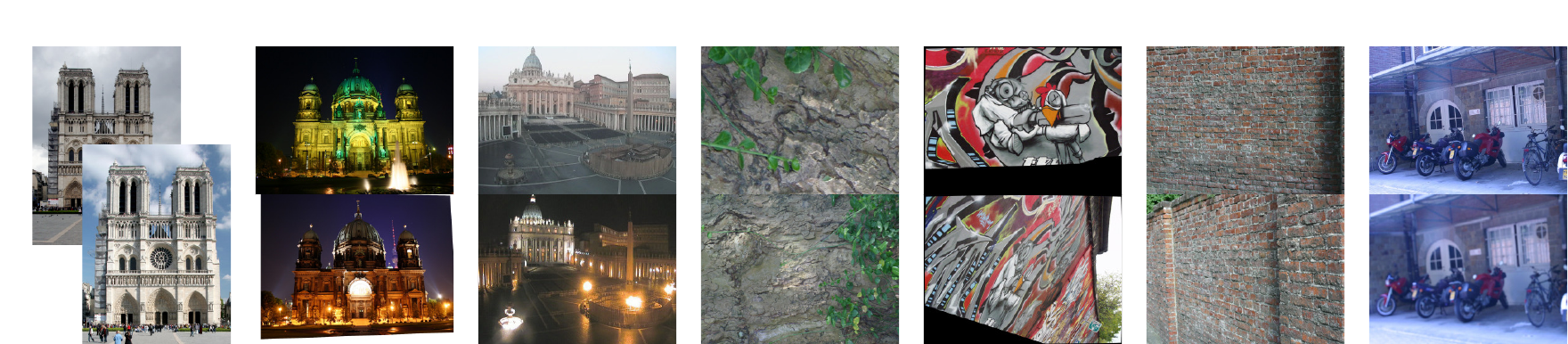
\end{center}
\caption{First row: descriptors computed using authors' implementation, second row - descriptors computed on photometrically normalized patches (mean = 0.5, var = 0.2) patches as done in SIFT. Third row: top 5 complementary pairs of descriptors (photometrically normalized). The numbers in legend are mean average precision. Bottom row: examples of the image pairs from each subset. Note that axis scales differs in each column, i.e. for different WxBS problems.}
\label{fig:desc_matching_perf}
\vspace{-1em}
\end{figure*}
Floating point descriptors have been compared using $L_2$ distance, binary using Hamming distance. The Recall-Precision curves are shown in Figure~\ref{fig:desc_matching_perf}. The second-nearest distance ratio is used to parameter the curve for floating point descriptors, the Hamming distance for binary ones. 

Note that most of the descriptors gain significantly from photometric normalization, cf. the first two rows of
Figure~\ref{fig:desc_matching_perf}. The published implementations are clearly sensitivite to contrast variations. 

The results hows that gradient-histogram based SIFT and its variants including DAISY are the best performing descriptors by a big margin in the presence of any (geometric, illumination, etc) nuisance factors
despite the fact that some of the competitors -- LIOP, MROGH -- have been specifically designed to deal with illumination changes. The second best descriptor is -- surprisingly -- the patch with contrast-$L_2$-normalized  pixels, which beats all other descriptors. It has huge memory footprint -- 1681 floats, but the affine-photo-$L_2$-normed grayscale pixel intensities are a strong descriptor baseline.

Most of descriptors, despite their different underlying assumptions and algorithmic structure, successfully match almost the same patches (see third row in Figure~\ref{fig:desc_matching_perf}) -- and the most complementary descriptor to the leading rSIFT is its gradient-reversal-insensitive version -- hrSIFT.

The results confirming the domination of SIFT-based methods are in agreement with \cite{Stylianou2015} and \cite{Fernando2014} despite the fact that they adopted a rather different evaluation methodology. However, we could not confirm clear superiority of the SSIM over SymFeat descriptors, which could be explained by the fact that the SSIM descriptor was designed for use only with the SSIM detector.

\noindent
{\bf Detectors evaluation. } The following detectors are compared:  MSER~\cite{Matas2002}, DoG~\cite{Lowe2004}, Hessian-Affine~\cite{Mikolajczyk2004} (implementation~\cite{Perdoch2009}), FOCI~\cite{Zitnick2011}, IIDOG~\cite{Vonikakis2013}, WADE~\cite{Salti2013}, W$\alpha$SH~\cite{Varytimidis2012}, SURF~\cite{Bay2006}, SFOP~\cite{foerstner09}, AKAZE\cite{Alcantarilla2013}.
%
We focus on getting a reliable answer to the  "match/non-match" question in real image pairs. Therefore the performance criterion is the number of successfully matched pairs using the best combination of descriptors (see Section {\bf Descriptors evaluation} ) -- ~rSIFT and hrSIFT. Matching is done as in Algorithm~\ref{alg:wxbs-m} except that no view synthesis is performed. Image pairs are considered matched if $\geq$15 correct inliers to a homography are found. Since the Lost-in-past dataset contains 2300 matchable image pairs, which is unfeasible for direct matching, we have selected a subset of 172 medium-challenging image pairs. Other datasets are used fully.

\noindent
{\bf Adaptive threshold of the detector response.}
One of the main problems in matching of day to night and infrared images is the low number of detected features. The problem is acute in dark low contrast images  in the \textsc{WgsBS} and MMS~\cite{Aguilera2012} datasets. 
A possible approach addressing the problem is iiDoG~\cite{Vonikakis2013} where the difference of Gaussians is normalized by sum of Gaussians. It works well, but cannot be easily applied for other types of detectors, i.e. MSER. 

\begin{table}[tb]
\centering
\caption{Detector evaluation results. The number of matched image pairs (left) and the average running time (right). The FOCI detector is run through MS Windows simulator wine, the time includes a big overhead.}
\label{tab:detector-results}
\scriptsize
\setlength{\tabcolsep}{.25em}
\vspace{0.5em}
\begin{tabular}{|l|rr|rr|rr|rr|rr|rr|rr|rr|rr|rr|rr|rr|}
\hline

\centering{Alg.}
& \multicolumn{2}{c|}{EF}
& \multicolumn{2}{c|}{EVD}
& \multicolumn{2}{c|}{MMS}
& \multicolumn{2}{c|}{\textsc{WgaBS}}
& \multicolumn{2}{c|}{\textsc{WgalBS}}
& \multicolumn{2}{c|}{\textsc{WglBS}}
& \multicolumn{2}{c|}{\textsc{WgsBS}}
& \multicolumn{2}{c|}{\textsc{WlaBS}}
& \multicolumn{2}{c|}{Past}
& \multicolumn{2}{c|}{OxAff}
& \multicolumn{2}{c|}{SymB}
& \multicolumn{2}{c|}{GDB}
\\
 & \shortstack{\# \\33}& \shortstack{time \\ $[s]$}
& \shortstack{\# \\ 15}& \shortstack{time \\ $[s]$}
& \shortstack{\# \\100}& \shortstack{time \\ $[s]$}
& \shortstack{\# \\ 5}& \shortstack{time \\ $[s]$}
& \shortstack{\# \\8}& \shortstack{time \\ $[s]$}
& \shortstack{\# \\ 9}& \shortstack{time \\ $[s]$}
& \shortstack{\# \\ 5}& \shortstack{time \\ $[s]$}
& \shortstack{\#\\ 4}& \shortstack{time \\ $[s]$}
& \shortstack{\#\\172}& \shortstack{time \\ $[s]$}
& \shortstack{\#\\40}& \shortstack{time \\ $[s]$}
& \shortstack{\#\\46}& \shortstack{time \\ $[s]$}
& \shortstack{\# \\22}& \shortstack{time \\ $[s]$}
\\
\hline
\multicolumn{25}{|c|}{Threshold adaptation}\\
\hline
MSER & \ccaEF{16} & 1.4 & \ccaEVD{3} & 1.4 & \ccaMMS{1} & 0.3 & \ccaWGABS{0} & 2.0 & \ccaWGALBS{0} & 1.3 & \ccaWGLBS{0} & 1.3 & \ccaWGSBS{0} & 0.8 & \ccaWLABS{1} & 1.2 & \ccalostinpast{8} & 1.3 & \ccaoxford{40} & 3.5 & \ccasnavely{23} & 2.4 & \ccastewart{9} & 2.4\\
AdMSER & \ccaEF{25} & 3.4 & \ccaEVD{8} & 4.0 & \ccaMMS{6} & 1.0 & \ccaWGABS{0} & 4.0 & \ccaWGALBS{0} & 3.2 & \ccaWGLBS{0} & 3.3 & \ccaWGSBS{0} & 1.4 & \ccaWLABS{1} & 2.6 & \ccalostinpast{11} & 2.9 & \ccaoxford{40} & 5.7 & \ccasnavely{26} & 4.6 & \ccastewart{13} & 6.9\\
DoG & \ccaEF{29} & 2.3 & \ccaEVD{0} & 2.8 & \ccaMMS{10} & 0.8 & \ccaWGABS{0} & 2.7 & \ccaWGALBS{0} & 2.3 & \ccaWGLBS{0} & 2.1 & \ccaWGSBS{0} & 1.0 & \ccaWLABS{1} & 2.4 & \ccalostinpast{13} & 2.0 & \ccaoxford{38} & 4.8 & \ccasnavely{29} & 2.7 & \ccastewart{12} & 4.7\\
iiDoG & \ccaEF{29} & 3.1 & \ccaEVD{0} & 3.0 & \ccaMMS{11} & 1.2 & \ccaWGABS{0} & 3.2 & \ccaWGALBS{0} & 2.9 & \ccaWGLBS{0} & 2.8 & \ccaWGSBS{0} & 1.2 & \ccaWLABS{1} & 2.5 & \ccalostinpast{13} & 2.2 & \ccaoxford{38} & 8.0 & \ccasnavely{29} & 2.9 & \ccastewart{12} & 6.1\\
AdDoG & \ccaEF{29} & 2.6 & \ccaEVD{0} & 3.4 & \ccaMMS{11} & 1.2 & \ccaWGABS{0} & 3.3 & \ccaWGALBS{0} & 3.0 & \ccaWGLBS{0} & 3.0 & \ccaWGSBS{0} & 1.5 & \ccaWLABS{1} & 2.7 & \ccalostinpast{13} & 2.7 & \ccaoxford{38} & 4.1 & \ccasnavely{30} & 3.0 & \ccastewart{12} & 4.8\\
HesAf & \ccaEF{32} & 4.6 & \ccaEVD{1} & 5.2 & \ccaMMS{15} & 1.2 & \ccaWGABS{0} & 5.5 & \ccaWGALBS{0} & 3.8 & \ccaWGLBS{0} & 4.2 & \ccaWGSBS{0} & 2.0 & \ccaWLABS{1} & 3.6 & \ccalostinpast{24} & 4.0 & \ccaoxford{40} & 11. & \ccasnavely{35} & 5.8 & \ccastewart{17} & 9.1\\
AdHesAf & \ccaEF{33} & 5.7 & \ccaEVD{2} & 7.6 & \ccaMMS{35} & 2.9 & \ccaWGABS{0} & 7.2 & \ccaWGALBS{1} & 6.5 & \ccaWGLBS{0} & 6.0 & \ccaWGSBS{0} & 3.2 & \ccaWLABS{1} & 4.9 & \ccalostinpast{25} & 5.4 & \ccaoxford{40} & 10. & \ccasnavely{35} & 7.2 & \ccastewart{18} & 13.\\
\hline
\multicolumn{25}{|c|}{Other detectors}\\
\hline
W$\alpha$SH & \ccaEF{0} & 1.8 & \ccaEVD{0} & 5.4 & \ccaMMS{0} & 0.6 & \ccaWGABS{0} & 2.8 & \ccaWGALBS{0} & 2.5 & \ccaWGLBS{0} & 1.4 & \ccaWGSBS{0} & 1.8 & \ccaWLABS{0} & 1.2 & \ccalostinpast{0} & 1.9 & \ccaoxford{24} & 4.1 & \ccasnavely{3} & 2.8 & \ccastewart{3} & 6.9\\
ORB & \ccaEF{3} & 4.1 & \ccaEVD{0} & 3.6 & \ccaMMS{1} & 0.8 & \ccaWGABS{0} & 2.8 & \ccaWGALBS{0} & 2.7 & \ccaWGLBS{0} & 3.6 & \ccaWGSBS{0} & 1.6 & \ccaWLABS{0} & 2.8 & \ccalostinpast{1} & 2.3 & \ccaoxford{28} & 8.7 & \ccasnavely{5} & 3.0 & \ccastewart{3} & 6.1\\
SURF & \ccaEF{27} & 2.3 & \ccaEVD{0} & 2.4 & \ccaMMS{7} & 1.0 & \ccaWGABS{0} & 2.5 & \ccaWGALBS{0} & 1.9 & \ccaWGLBS{0} & 2.1 & \ccaWGSBS{0} & 0.9 & \ccaWLABS{1} & 1.4 & \ccalostinpast{10} & 1.9 & \ccaoxford{38} & 5.8 & \ccasnavely{31} & 2.9 & \ccastewart{15} & 4.0\\
AKAZE & \ccaEF{28} & 4.3 & \ccaEVD{0} & 3.6 & \ccaMMS{10} & 0.8 & \ccaWGABS{1} & 4.7 & \ccaWGALBS{0} & 3.4 & \ccaWGLBS{0} & 4.0 & \ccaWGSBS{0} & 1.3 & \ccaWLABS{1} & 2.7 & \ccalostinpast{25} & 3.6 & \ccaoxford{38} & 13. & \ccasnavely{35} & 5.6 & \ccastewart{17} & 6.4\\
FOCI & \ccaEF{29} & 12. & \ccaEVD{0} & 39. & \ccaMMS{14} & 11. & \ccaWGABS{1} & 32. & \ccaWGALBS{0} & 29. & \ccaWGLBS{0} & 29. & \ccaWGSBS{0} & 20. & \ccaWLABS{1} & 29. & \ccalostinpast{21} & 13. & \ccaoxford{38} & 35. & \ccasnavely{35} & 27. & \ccastewart{17} & 45.\\
SFOP & \ccaEF{25} & 11. & \ccaEVD{0} & 16. & \ccaMMS{12} & 4.7 & \ccaWGABS{0} & 12. & \ccaWGALBS{0} & 10. & \ccaWGLBS{0} & 10. & \ccaWGSBS{0} & 9.2 & \ccaWLABS{0} & 7.5 & \ccalostinpast{11} & 12. & \ccaoxford{36} & 15. & \ccasnavely{24} & 11. & \ccastewart{8} & 17.\\
WADE & \ccaEF{16} & 14. & \ccaEVD{0} & 20. & \ccaMMS{0} & 3.4 & \ccaWGABS{0} & 58. & \ccaWGALBS{0} & 11. & \ccaWGLBS{0} & 14. & \ccaWGSBS{0} & 7.9 & \ccaWLABS{1} & 8.3 & \ccalostinpast{20} & 23. & \ccaoxford{34} & 60. & \ccasnavely{34} & 46. & \ccastewart{13} & 77.\\
\hline
\multicolumn{25}{|c|}{State-of-art matchers}\\
\hline
ASIFT   &\ccaEF{23}&27.& \ccaEVD{5}          &12. & \ccaMMS{18}&3.2& 0 & 52.& 0&32.&0&35.&0&12.&\ccaWLABS{1}&30.&\ccalostinpast{62}&32.&\ccaoxford{40} &102& \ccasnavely{27}& 14.&  \ccastewart{15}&  41.\\
MODS & \ccaEF{33} & 4.8 & \cellcolor{black}\color{white}{15} & 11. & \ccaMMS{27} & 11. & \ccaWGABS{2} & 41. & \ccaWGALBS{2} & 31. & \ccaWGLBS{1} & 46. & \ccaWGSBS{0} & 17. & \ccaWLABS{1} & 26. & \ccalostinpast{94} & 27. & \ccaoxford{40} & 3.4 & \ccasnavely{42} & 18. & \ccastewart{18} & 11.\\
DBstrap &\ccaEF{31}&26.& 0          &18. &\ccaMMS{79}&9.3& 0&11.&0&13.&0&13.&0&4.7&0&15.& \ccalostinpast{16} & 28.& \ccaoxford{36}  &24. &  \ccasnavely{38}&    21.& \ccastewart{16} & 17.\\
\hline
\multicolumn{25}{|c|}{Proposed matcher}\\
\hline
WXBS-M & \ccaEF{33} & 4.7 & \cellcolor{black}\color{white}{15} & 14. & \ccaMMS{82} & 12. & \ccaWGABS{3} & 40. & \ccaWGALBS{3} & 63. & \ccaWGLBS{3} & 61. & \ccaWGSBS{0} & 26. & \ccaWLABS{3} & 28. & \cellcolor{black}\color{white}{107} & 42. & \ccaoxford{40} & 5.1 & \ccasnavely{43} & 18. &  \cellcolor{black}\color{white}{22}& 12.\\
\hline

\end{tabular}
\end{table}

Instead, we propose to use the following adaptive thresholding for all feature detectors. 
First, all local extrema of the response function are detected (i.e. no thresholding takes place). Next, the detected features are sorted according to the response magnitude. If the number of detected features with response magnitude $\geq$ $\Theta$ is greater than a given threshold $R_{\text{min}}$, these are  output and the algorithm terminates (this is the standard approach).
If there is not enough features above the threshold, top $R_{\text{min}}$ features our output. 

\noindent
{\bf Discussion and results.}
The performance of the proposed WxBS-M matcher is compared
with it state-of-art matchers: ASIFT~\cite{Morel2009}, Dual Bootstrap (DBstrap)~\cite{Yang2007} and MODS~\cite{Mishkin2015} on various WxBS problems.

The results are summarized in Table~\ref{tab:detector-results}. Note that the state-of-the-art matchers were not able to match almost any image pair which combines more nuisance factors. The proposed \textsc{WxBS-M} matcher shows much better performance, but still is not able to solve even half of the new dataset pairs.

Results in  Table~\ref{tab:detector-results} confirm that the proposed adaptive thresholding strategy works as well
as, or even better, than iiDoG for DoG, but it is 1.5 times faster. It also significantly improves results of the MSER and Hessian-Affine, even when main the nuisance is in the viewing geometry (EVD dataset).

\vspace{-1em}
\section{Conclusions}
\label{sec:conclusions}
\vspace{-0.5em}
We have presented a new problem -- the wide multiple baseline stereo (\textsc{WxBS}) -- which considers matching of images that simultaneously differ in more than one image acquisition factor such as viewpoint, illumination, sensor type or where object appearance changes significantly, e.g. over time. A new dataset with the ground truth for evaluation of matching algorithms has been introduced and will be made public.

We have extensively tested a large set of popular and recent detectors and descriptors and show than the combination of RootSIFT and HalfRootSIFT as descriptors with MSER and Hessian-Affine detectors works best for many different nuisance factors. 
We show that simple adaptive thresholding improves Hessian-Affine, DoG, MSER (and possibly other) detectors and allows to use them on infrared and low contrast images.

A novel matching algorithm for addressing the WxBS problem has been introduced. We have shown experimentally that the \textsc{WxBS-M} matcher dominantes the state-of-the-art methods both on both the new and existing datasets.

\begin{spacing}{0.8}
\small
\bibliography{bmvc_review}
\end{spacing}
\end{document}